\documentclass[journal]{IEEEtran}

\usepackage{graphicx}
\usepackage{bm}
\usepackage{tikz}
\usepackage{array}
\usepackage{arydshln}
\usepackage{subfigure}
\usepackage{authblk}
\usepackage{multirow}
\usepackage{url}
\usepackage{rotating}
\usepackage{makecell}

\usetikzlibrary{arrows,positioning}
\usetikzlibrary{plotmarks}
\usetikzlibrary{calc}
\usetikzlibrary{shapes.geometric}

\ifCLASSINFOpdf
\else
\fi
\newcommand{\resneti}{\mathcal{I}_i^{(rn)}}
\newcommand{\resnetset}{\bm{I}^{(rn)}}
\newcommand{\lrset}{\bm{I}^{(l)}}
\newcommand{\lri}{\mathcal{I}_i^{(l)}}
\newcommand{\hr}{\mathcal{I}^{(h)}}
\newcommand{\hrsrr}{\mathcal{I}^{(sr)}}

\newcommand{\algname}{EvoNet}

\graphicspath{{Figures/}}

\begin{document}
\title{Deep Learning for Multiple-Image Super-Resolution}
%
%
%

\author{Michal Kawulok,~\IEEEmembership{Member,~IEEE}, 
        Pawel Benecki,
        Szymon Piechaczek,
        Krzysztof Hrynczenko,
        Daniel Kostrzewa,
        and Jakub Nalepa,~\IEEEmembership{Member,~IEEE}
\thanks{The reported work was funded by European Space Agency (SuperDeep project, realized by Future Processing). MK and JN were partially supported by the National Science Centre under Grant DEC-2017/25/B/ST6/00474. PB and DK were supported by the Silesian University of Technology, Poland, funds no. BKM-509/RAu2/2017.
}
\thanks{M.~Kawulok, P.~Benecki, S.~Piechaczek, K.~Hrynczenko, D.~Kostrzewa, and J.~Nalepa are with Future Processing, Gliwice, Poland and with Silesian University of Technology, Gliwice, Poland (e-mail: michal.kawulok@ieee.org).}
}
\maketitle

\begin{abstract}
Super-resolution reconstruction (SRR) is a process aimed at enhancing spatial resolution of images, either from a single observation, based on the learned relation between low and high resolution, or from multiple images presenting the same scene. SRR is particularly valuable, if it is infeasible to acquire images at desired resolution, but many images of the same scene are available at lower resolution---this is inherent to a variety of remote sensing scenarios. Recently, we have witnessed substantial improvement in single-image SRR attributed to the use of deep neural networks for learning the relation between low and high resolution. Importantly, deep learning has not been exploited for multiple-image SRR, which benefits from information fusion and in general allows for achieving higher reconstruction accuracy. In this letter, we introduce a new method which combines the advantages of multiple-image fusion with learning the low-to-high resolution mapping using deep networks. The reported experimental results indicate that our algorithm outperforms the state-of-the-art SRR methods, including these that operate from a single image, as well as those that perform multiple-image fusion.

\end{abstract}


%
\IEEEpeerreviewmaketitle

\section{Introduction}

\emph{Super-resolution reconstruction} (SRR) is aimed at generating a \emph{high-resolution} (HR) image from a single or multiple \emph{low-resolution} (LR) observations. In many cases, the SRR algorithms are the only possibility to obtain images of sufficient spatial resolution, as HR data may not be available due to high acquisition costs or sensor limitations. Such situations are an inherent problem to remote sensing, in particular concerning satellite imaging for Earth observation purposes.

The existing approaches towards SRR can be categorized into single-image and multiple-image methods. The former consist in learning the LR-HR relation from a large number of examples. This relation allows us to reconstruct an HR image from an LR scene (unseen during training). Multiple-image SRR is based on information fusion, which benefits from the differences (mainly subpixel shifts) between LR images---in general, these approaches allow for more accurate reconstruction  than single-image SRR, as they combine more data extracted from the analyzed scene. 
The recent advancements in \emph{deep learning}, especially in deep \emph{convolutional neural networks} (CNNs), have greatly improved single-image SRR, however it is worth noting that correct fusion of multiple LR images still offers higher reconstruction accuracy. Despite that, to the best of our knowledge, deep learning has not been employed for multiple-image SRR. 

In this letter, our contribution lies in combining the advantages of single-image SRR based on deep learning with the benefits of information fusion offered by multiple-image reconstruction (Section~\ref{sec:overview} presents the related work). We introduce \algname~(Section~\ref{sec:method}), which employs a \emph{deep residual network}, more specifically ResNet~\cite{Ledig2017}, to enhance the capabilities of \emph{evolutionary imaging model} (EvoIM)~\cite{Kawulok2018Gecco} for multiple-image SRR. The results of our extensive experimental validation (Section~\ref{sec:exp}) focused on satellite imaging are highly encouraging and they  show that \algname~renders qualitatively and quantitatively better outcome than the state-of-the-art techniques for single-image and multiple-image SRR.


\section{Related Work} \label{sec:overview}

In this section, we outline the state of the art in multiple-image SRR (Section~\ref{sec:multi}), and we present the recent advancements in using deep learning for SRR (Section~\ref{sec:deep}).

\subsection{Multiple-image super-resolution reconstruction} \label{sec:multi}

Existing techniques for multiple-image SRR are based on the premise that each LR observation $\lri$ in a set $\lrset=\left\{\lri:i \in \left\{1,2,\cdots,N\right\}\right\}$ has been derived from an original HR image $\hr$, degraded using an assumed \emph{imaging model} (IM) that usually includes image warping, blurring, decimation and contamination with the noise. The reconstruction consists in reversing that degradation process, which requires solving an ill-posed optimization problem, therefore most SRR techniques employ some regularization to provide spatial smoothness of the reconstructed HR image $\hrsrr$. In one of the earliest approaches, Irani and Peleg performed SRR relying on image registration (hence reducing the IM to subpixel shifts)~\cite{Irani1991}. A hierarchical subpixel displacement estimation was combined with the Bayesian reconstruction in the \emph{gradient projection algorithm} (GPA)~\cite{Schultz1996}. 
Another popular optimization technique applied here is the \emph{projection onto convex sets}~\cite{Akgun2005}, which consists in updating the HR target image iteratively based on the error measured between $\mathcal{I}^{(l)}$ and a downsampled version of the reconstruction outcome $\hrsrr$, degraded using the assumed IM. Farsiu et al. introduced \emph{fast and robust super-resolution} (FRSR)~\cite{Farsiu2004} based on \emph{maximum likelihood} estimation coupled with simplified regularization---importantly, the error is measured in the HR coordinates, thus avoiding the expensive scaling operation. Among other methods, adaptive Wiener filter~\cite{Hardie2007} and random Markov fields~\cite{LiJia2008} were used to specify the IM. Zhu et al. proposed \emph{adaptive detail enhancement} (SR-ADE)~\cite{Zhu2016} for reconstructing satellite images---a bilateral filter is employed to decompose the input images and amplify the high-frequency detail information.

Recently, we proposed the EvoIM method~\cite{Kawulok2018Gecco,Kawulok2018EvoStar}, which employs a genetic algorithm to optimize the hyper-parameters that control the IM used in FRSR~\cite{Farsiu2004}, and to evolve the convolution kernels instead of the Gaussian blur used in FRSR. We showed that the reconstruction process can be effectively adapted to different imaging conditions---in particular, we used Sentinel-2 images at original resolution as LR inputs, and compared the reconstruction outcome with SPOT images presenting the same region.

\begin{figure*}[!th]
\centering
\resizebox{\textwidth}{!}{

\newcommand{\largefig}{120pt}
\newcommand{\mediumfig}{60pt}
\newcommand{\inputfig}{30pt}

\newcommand{\resnettext}{Reconstruction with ResNet}
\newcommand{\registrationtext}{Image registration}
\newcommand{\shifttext}{Shifts $(s^x_i,s^y_i)$}
\newcommand{\evoimtext}{EvoIM iterative image filtering}
\newcommand{\groundtruthtext}{$M$ ground-truth high-resolution images $\mathcal{I}^{(h)}$}

\newcommand{\blocksep}{45pt}
\newcommand{\multiblock}{5pt}
\newcommand{\arrowsep}{15pt}

\begin{tikzpicture}[scale=0.8,
        image/.style={inner sep=0pt,draw=white,very thick},
        whitefont/.style={text=white,font=\footnotesize},
        input/.style={rectangle,draw=black,fill=gray!40,inner sep=5pt,minimum height=27pt,minimum width=40pt,text width=50pt,text badly centered,font=\small,thick},
        output/.style={rectangle,draw=black,fill=red!20,inner sep=5pt,minimum height=32pt,minimum width=40pt,text width=65pt,text badly centered,font=\footnotesize,thick},
        ga/.style={rectangle,draw=black,fill=red!20,inner sep=5pt,minimum height=32pt,minimum width=40pt,text width=100pt,text badly centered,font=\footnotesize,thick,rounded
        corners=16pt},
        algstep/.style={rectangle,draw=black,fill=blue!20,inner sep=5pt,minimum height=27pt,minimum width=40pt,text width=60pt,text badly centered,font=\small,thick,rounded
        corners=8pt},
        plain/.style={rectangle,text width=40pt,text badly centered,font=\small},
        myarrow/.style={thick}]

\node (img_input_a) [image] {\includegraphics[width=\inputfig]{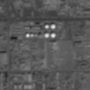}};
\node (img_input_b) [image, above left=\multiblock of img_input_a.north, anchor=north] {\includegraphics[width=\inputfig]{example_flowchart_input.png}};
\node (img_input_c) [image, above left=\multiblock of img_input_b.north, anchor=north] {\includegraphics[width=\inputfig]{example_flowchart_input.png}};
\node (img_input_d) [image, above left=\multiblock of img_input_c.north, anchor=north] {\includegraphics[width=\inputfig]{example_flowchart_input.png}};

\node (resnet_a) [algstep, right=\blocksep of img_input_a] {\resnettext};
\node (resnet_b) [algstep, right=\blocksep of img_input_b] {\resnettext};
\node (resnet_c) [algstep, right=\blocksep of img_input_c] {\resnettext};
\node (resnet_d) [algstep, right=\blocksep of img_input_d] {\resnettext};

\node (img_resnet_a) [image, right=\blocksep of resnet_a] {\includegraphics[width=\mediumfig]{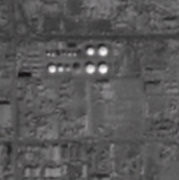}};
\node (img_resnet_b) [image, above left=\multiblock of img_resnet_a.north, anchor=north] {\includegraphics[width=\mediumfig]{example_flowchart_resnet.png}};
\node (img_resnet_c) [image, above left=\multiblock of img_resnet_b.north, anchor=north] {\includegraphics[width=\mediumfig]{example_flowchart_resnet.png}};
\node (img_resnet_d) [image, above left=\multiblock of img_resnet_c.north, anchor=north] {\includegraphics[width=\mediumfig]{example_flowchart_resnet.png}};

\node (registration) [algstep, below=\blocksep of resnet_d.south west, anchor=north west, text width=68, minimum height=35] {\registrationtext};

\node (shift_a) [input, below=\blocksep of img_resnet_a,anchor=center] {\shifttext};
 \draw[-,myarrow,white] ($(shift_a.north)$) -- ($(img_resnet_a.south)$);
 \draw[-,myarrow,dashed] ($(shift_a.north)$) -- ($(img_resnet_a.south)$);
\node (shift_b) [input, below=\blocksep of img_resnet_b,anchor=center] {\shifttext};
 \draw[-,myarrow,white] ($(shift_b.north)$) -- ($(img_resnet_b.south)$);
 \draw[-,myarrow,dashed] ($(shift_b.north)$) -- ($(img_resnet_b.south)$);
\node (shift_c) [input, below=\blocksep of img_resnet_c,anchor=center] {\shifttext};
\draw[-,myarrow,white] ($(shift_c.north)$) -- ($(img_resnet_c.south)$);
 \draw[-,myarrow,dashed] ($(shift_c.north)$) -- ($(img_resnet_c.south)$);
\node (shift_d) [input, below=\blocksep of img_resnet_d,anchor=center] {\shifttext};
 \draw[-,myarrow,white] ($(shift_d.north)$) -- ($(img_resnet_d.south)$);
 \draw[-,myarrow,dashed] ($(shift_d.north)$) -- ($(img_resnet_d.south)$);

\node (img_initial_large) [image, right=\blocksep of img_resnet_b.north east,anchor=north west] {\includegraphics[width=\largefig]{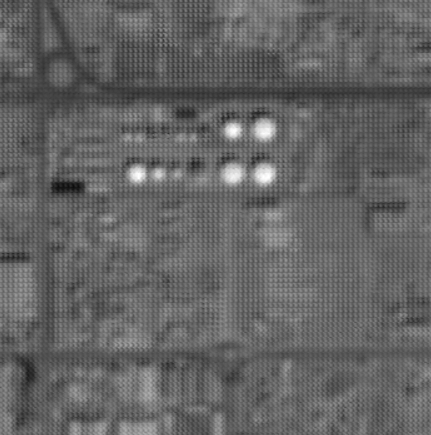}};

\node (evoim) [algstep, right=30pt of img_initial_large, text width=40pt] {\evoimtext};

\node (img_final) [image, right=30pt of evoim] {\includegraphics[width=\largefig]{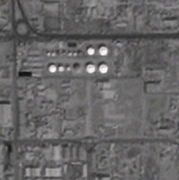}};

 \draw[->,myarrow] ($(img_input_a.east) + (\arrowsep,0)$) -- ($(resnet_a.west) + (-\arrowsep,0)$);
 \draw[->,myarrow] ($(img_input_b.east) + (\arrowsep,0)$) -- ($(resnet_b.west) + (-\arrowsep,0)$);
 \draw[->,myarrow] ($(img_input_c.east) + (\arrowsep,0)$) -- ($(resnet_c.west) + (-\arrowsep,0)$);
 \draw[->,myarrow] ($(img_input_d.east) + (\arrowsep,0)$) -- ($(resnet_d.west) + (-\arrowsep,0)$);

 \draw[->,myarrow] ($(resnet_a.east) + (\arrowsep,0)$) -- ($(img_resnet_a.west) + (-\arrowsep,0)$);
 \draw[->,myarrow] ($(resnet_b.east) + (\arrowsep,0)$) -- ($(img_resnet_b.west) + (-\arrowsep,0)$);
 \draw[->,myarrow] ($(resnet_c.east) + (\arrowsep,0)$) -- ($(img_resnet_c.west) + (-\arrowsep,0)$);
 \draw[->,myarrow] ($(resnet_d.east) + (\arrowsep,0)$) -- ($(img_resnet_d.west) + (-\arrowsep,0)$);

 \draw[->,myarrow] ($(img_input_a.south) + (0,-\arrowsep)$)  .. controls +(-90:1.6) and +(-180:1.2) .. ($(registration.west) + (-4pt,0)$);
 \draw[->,myarrow] ($(img_input_b.south) + (0,-\arrowsep)$)  .. controls +(-90:1.7) and +(-180:1.2) .. ($(registration.west) + (-4pt,0)$);
 \draw[->,myarrow] ($(img_input_c.south) + (0,-\arrowsep)$)  .. controls +(-90:1.8) and +(-180:1.2) .. ($(registration.west) + (-4pt,0)$);
 \draw[->,myarrow] ($(img_input_d.south) + (0,-\arrowsep)$)  .. controls +(-90:1.9) and +(-180:1.2) .. ($(registration.west) + (-4pt,0)$);

 \draw[->,myarrow] ($(registration.east) + (4pt,0)$)  .. controls +(0:0.6) and +(-180:0.6) .. ($(shift_a.west) + (-\arrowsep,0)$);
 \draw[->,myarrow] ($(registration.east) + (4pt,0)$)  .. controls +(0:0.6) and +(-180:0.6) .. ($(shift_b.west) + (-\arrowsep,0)$);
 \draw[->,myarrow] ($(registration.east) + (4pt,0)$)  .. controls +(0:0.6) and +(-180:0.6) .. ($(shift_c.west) + (-\arrowsep,0)$);
 \draw[->,myarrow] ($(registration.east) + (4pt,0)$)  .. controls +(0:0.6) and +(-180:0.6) .. ($(shift_d.west) + (-\arrowsep,0)$);

 \draw[->,myarrow] ($(shift_a.east) + (\arrowsep,0)$)  .. controls +(0:0.6) and +(-180:1) .. ($(img_initial_large.west) + (-4pt,0)$);
 \draw[->,myarrow] ($(shift_b.east) + (\arrowsep,0)$)  .. controls +(0:0.7) and +(-180:1) .. ($(img_initial_large.west) + (-4pt,0)$);
 \draw[->,myarrow] ($(shift_c.east) + (\arrowsep,0)$)  .. controls +(0:0.8) and +(-180:1) .. ($(img_initial_large.west) + (-4pt,0)$);
 \draw[->,myarrow] ($(shift_d.east) + (\arrowsep,0)$)  .. controls +(0:0.9) and +(-180:1) .. ($(img_initial_large.west) + (-4pt,0)$);

 \draw[->,myarrow] ($(img_resnet_a.east) + (\arrowsep,0)$)  .. controls +(0:0.4) and +(-180:1) .. ($(img_initial_large.west) + (-4pt,0)$);
 \draw[->,myarrow] ($(img_resnet_b.east) + (\arrowsep,0)$)  .. controls +(0:0.7) and +(-180:1) .. ($(img_initial_large.west) + (-4pt,0)$);
 \draw[->,myarrow] ($(img_resnet_c.east) + (\arrowsep,0)$)  .. controls +(0:1) and +(-180:1) .. ($(img_initial_large.west) + (-4pt,0)$);
 \draw[->,myarrow] ($(img_resnet_d.east) + (\arrowsep,0)$)  .. controls +(0:1.3) and +(-180:1) .. ($(img_initial_large.west) + (-4pt,0)$);

 \draw[->,myarrow] ($(img_initial_large.east) + (4pt,0)$) -- ($(evoim.west) + (-4pt,0)$);
 \draw[->,myarrow] ($(evoim.east) + (4pt,0)$) -- ($(img_final.west) + (-4pt,0)$);

 \node [plain, above=1pt of img_input_d,text width=60pt] {Input LR images $\left( \lrset \right)$};
 \node [plain,above=1pt of img_resnet_d, text width=120pt] {ResNet outcomes $\left( \resnetset \right)$};
 \node [plain,above=1pt of img_initial_large, text width=120pt] {Shift-and-add fusion $\left(\mathcal{X}_0\right)$};
 \node [plain,above=1pt of img_final.north, text width=120pt, anchor=south] {Final SRR outcome $\left( \hrsrr \right)$};

\end{tikzpicture}
}
  \caption{Flowchart of the proposed EvoNet algorithm---a set of $N$ input images ($\lrset$) is processed with ResNet, integrated into $\mathcal{X}_0$ based on the shifts computed from $\lrset$, and the final reconstruction outcome $\hrsrr$ is obtained using EvoIM.}
  \label{fig:flowchart}
\end{figure*}

\subsection{Deep learning for single-image super-resolution} \label{sec:deep}

Inspired by earlier approaches based on sparse coding~\cite{ChavezRoman2014}, Dong et al. proposed \emph{super-resolution CNN} (SRCNN)~\cite{Dong2014}, followed by its faster version (FSRCNN)~\cite{Dong2016a}, for learning the LR-to-HR mapping from a number of LR--HR image pairs. Despite relatively simple architecture, SRCNN outperforms the state-of-the-art example-based methods. Liebel and Korner have successfully trained SRCNN with Sentinel-2 images, improving its capacities of enhancing satellite data~\cite{Liebel2016}. The same architecture was used to improve spatial resolution of sea surface temperature maps~\cite{Ducournau2016}. Kim et al. addressed certain limitations of SRCNN with a \emph{very deep super-resolution} network~\cite{Kim2016} which can be efficiently trained relying on fast residual learning. The domain expertise was exploited using a \emph{sparse coding network}~\cite{Liu2016}, which achieves high training speed and model compactness. Lai et al. proposed deep Laplacian pyramid networks with progressive upsampling~\cite{Lai2017}, aimed at achieving high processing speed. Recently, \emph{generative adversarial networks} (GANs) are being actively explored for SRR~\cite{Ledig2017}. GANs are composed of a generator (ResNet in~\cite{Ledig2017}), trained to perform SRR, whose outcome is classified by a discriminator, learned to distinguish between the images reconstructed by the generator and the real HR images (used for reference). In this way, the generator is promoted for generating images that are hard to distinguish from the real ones, thus it also learns avoiding the artifacts.

\section{The proposed \algname~algorithm} \label{sec:method}

A flowchart of the proposed method is presented in Fig.~\ref{fig:flowchart}. First of all, each of LR input images ($\lri$) is subject to single-image SRR using ResNet. This step produces a set of $N$ images $\resnetset=\{\resneti\}$, whose dimensions are $2\times$ larger than those of $\lri$. In parallel to that, the LR input set $\lrset$ undergoes image registration to determine subpixel shifts between the images. The obtained single-image SRR outcomes ($\resnetset$) alongside the subpixel shifts allow for composing the initial HR image $\mathcal{X}_0$ using the median shift-and-add method (the dimensions are increased again $2\times$, hence $4\times$ compared with $\lri$). Finally, $\mathcal{X}_0$ is subject to the iterative EvoIM process, which produces the final reconstruction outcome $\hrsrr$.

\subsection{Residual neural network applied to the input images}
Each LR image $\lri$ is independently enhanced using ResNet to obtain a higher-quality input data ($\resneti$) for further multiple-image fusion. For this purpose, we exploit the architecture described in~\cite{Ledig2017}, which is composed of 16 residual blocks with skip connections, and it is trained employing the \emph{mean square error} (MSE) as the loss function (during training, ResNet is guided to reduce MSE between each HR image and the reconstruction outcome obtained from the artificially-degraded HR image). For \algname, we modify the final layer, which determines the upscaling factor ($2\times$ in our case, compared with $4\times$ in~\cite{Ledig2017}).

\subsection{Multiple-image fusion}
The EvoIM process, which we employ for multiple-image fusion, consists in iterative filtering of an HR image $\mathcal{X}_0$, composed of registered LR inputs. In \algname, we register the original $\lri$ images, before they are processed with ResNet (the ResNet reconstruction does not introduce any information that may contribute to better assessment of the displacement values). As the dimensions of the ResNet outputs are $2\times$ larger than those of $\lri$, the computed shift values are multiplied by 2 to compose $\mathcal{X}_0$. Subsequently, EvoIM solves the optimization problem (analogously to the FRSR method). The update step $\Delta\mathcal{X} = \mathcal{X}_{n+1} - \mathcal{X}_n$ is computed as:
\begin{equation} \label {eq:step}
\footnotesize
\Delta\mathcal{X}= - \beta \left[ { \bm{B}' \bm{A}^T {\rm sgn}(\bm{A} \bm{B} \mathcal{X}_n - \bm{A} \mathcal{X}_0 )} + \lambda \frac{\delta U(\mathcal{X})}{ \delta \mathcal{X}} (\mathcal{X}_n) \right] \rm,
\end{equation}
where $\beta$ is a hyper-parameter that controls the update step, $\bm{A}$ is a diagonal matrix representing the number of the LR measurements that contributed to $\mathcal{X}_0$, $U(\mathcal{X})$ is the regularization term controlled with the $\lambda$ hyper-parameter, while $\bm{B}$ and $\bm{B}'$ are $5\times 5$ convolution kernels (in FRSR, $\bm{B}$ is the Gaussian blur and $\bm{B}'=\bm{B}^T$). The hyper-parameters alongside the convolution kernels are optimized during the EvoIM evolutionary training. Importantly, ResNet and EvoIM are trained separately before they are combined within the \algname~framework.

\begin{figure*}[!t]
\centering
\renewcommand{\tabcolsep}{0.2cm}
\scriptsize
\resizebox{\textwidth}{!}{
\newcommand{\w}{0.312}
\newcommand{\wenlarged}{0.312}

\newcommand{\file}{degraded_fragment2_10_hr.png}
\newcommand{\zoomfile}{degraded_fragment2_zoom_more_10_hr.png}

\begin{tikzpicture}[y=.018\columnwidth, x=.018\columnwidth, line/.style={color=blue!70}]
    \colorlet{line_color}{white}
    \colorlet{back_line_color}{red}
    \node (main) at (0,0) [inner sep=0] {\includegraphics[width=\w\columnwidth]{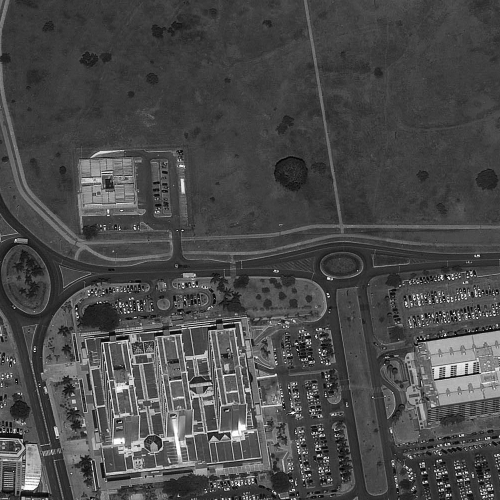}};
    \coordinate (enlarged) at ($(main.south west)+(0,0.5)$);
    \node (enlarged_img) at (enlarged) [anchor=north west,rectangle,draw=line_color,dashed,inner sep=0] {\includegraphics[width=\wenlarged\columnwidth]{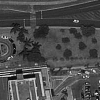}};
    \coordinate (enlarged_size) at ($(enlarged_img.north east)-(enlarged_img.south west)$);
    \coordinate (enlarged_width) at ($(enlarged_img.east)-(enlarged_img.west)$);
    \draw [color=back_line_color,thick] (enlarged_img.north west) rectangle (enlarged_img.south east);
    \draw [color=line_color,thick,dashed] (enlarged_img.north west) rectangle (enlarged_img.south east);

    \coordinate (enlarged_src_BL) at ($(main.north west)+55*(0.126,-0.222)$);
    \draw [color=back_line_color,thick] (enlarged_src_BL) rectangle ($(enlarged_src_BL)+0.21*(enlarged_size)$);
    \draw [color=line_color,dashed,thick] (enlarged_src_BL) rectangle ($(enlarged_src_BL)+0.21*(enlarged_size)$);

    \draw[color=back_line_color,thick,->] ($(enlarged_src_BL)+0.14*(enlarged_width)$) -- (enlarged_img);
    \draw[dashed,color=line_color,thick,->] ($(enlarged_src_BL)+0.14*(enlarged_width)$) -- (enlarged_img);

    \node [below=1 pt of enlarged_img] {\scriptsize High resolution};

\renewcommand{\file}{degraded_fragment2_10_lr_blur.png}
\renewcommand{\zoomfile}{degraded_fragment2_zoom_more_10_lr_blur.png}

    \node (main) at ($(main.north east) + (0.5,0)$) [anchor=north west, inner sep=0] {\includegraphics[width=\w\columnwidth]{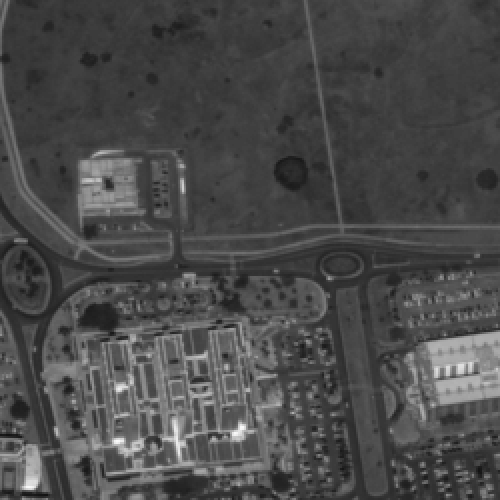}};
    \coordinate (enlarged) at ($(main.south west)+(0,0.5)$);
    \node (enlarged_img) at (enlarged) [anchor=north west,rectangle,draw=line_color,dashed,inner sep=0] {\includegraphics[width=\wenlarged\columnwidth]{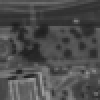}};
    \coordinate (enlarged_size) at ($(enlarged_img.north east)-(enlarged_img.south west)$);
    \coordinate (enlarged_width) at ($(enlarged_img.east)-(enlarged_img.west)$);
    \draw [color=back_line_color,thick] (enlarged_img.north west) rectangle (enlarged_img.south east);
    \draw [color=line_color,thick,dashed] (enlarged_img.north west) rectangle (enlarged_img.south east);

    \coordinate (enlarged_src_BL) at ($(main.north west)+55*(0.126,-0.222)$);
    \draw [color=back_line_color,thick] (enlarged_src_BL) rectangle ($(enlarged_src_BL)+0.21*(enlarged_size)$);
    \draw [color=line_color,dashed,thick] (enlarged_src_BL) rectangle ($(enlarged_src_BL)+0.21*(enlarged_size)$);

    \draw[color=back_line_color,thick,->] ($(enlarged_src_BL)+0.14*(enlarged_width)$) -- (enlarged_img);
    \draw[dashed,color=line_color,thick,->] ($(enlarged_src_BL)+0.14*(enlarged_width)$) -- (enlarged_img);

    \node [below=1 pt of enlarged_img] {\scriptsize Low resolution};

\renewcommand{\file}{degraded_fragment2_10_blur_resnet.png}
\renewcommand{\zoomfile}{degraded_fragment2_zoom_more_10_blur_resnet.png}

    \node (main) at ($(main.north east) + (0.5,0)$) [anchor=north west, inner sep=0] {\includegraphics[width=\w\columnwidth]{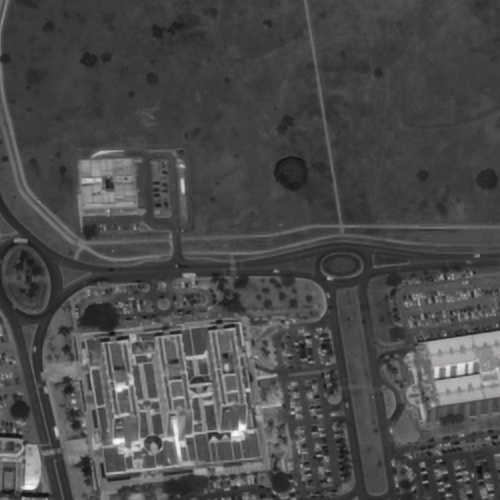}};
    \coordinate (enlarged) at ($(main.south west)+(0,0.5)$);
    \node (enlarged_img) at (enlarged) [anchor=north west,rectangle,draw=line_color,dashed,inner sep=0] {\includegraphics[width=\wenlarged\columnwidth]{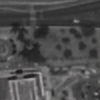}};
    \coordinate (enlarged_size) at ($(enlarged_img.north east)-(enlarged_img.south west)$);
    \coordinate (enlarged_width) at ($(enlarged_img.east)-(enlarged_img.west)$);
    \draw [color=back_line_color,thick] (enlarged_img.north west) rectangle (enlarged_img.south east);
    \draw [color=line_color,thick,dashed] (enlarged_img.north west) rectangle (enlarged_img.south east);

    \coordinate (enlarged_src_BL) at ($(main.north west)+55*(0.126,-0.222)$);
    \draw [color=back_line_color,thick] (enlarged_src_BL) rectangle ($(enlarged_src_BL)+0.21*(enlarged_size)$);
    \draw [color=line_color,dashed,thick] (enlarged_src_BL) rectangle ($(enlarged_src_BL)+0.21*(enlarged_size)$);

    \draw[color=back_line_color,thick,->] ($(enlarged_src_BL)+0.14*(enlarged_width)$) -- (enlarged_img);
    \draw[dashed,color=line_color,thick,->] ($(enlarged_src_BL)+0.14*(enlarged_width)$) -- (enlarged_img);

    \node [below=1 pt of enlarged_img] {\scriptsize ResNet~\cite{Ledig2017}};

\renewcommand{\file}{degraded_fragment2_10_blur_GASRR.png}
\renewcommand{\zoomfile}{degraded_fragment2_zoom_more_10_blur_GASRR.png}

    \node (main) at ($(main.north east) + (0.5,0)$) [anchor=north west, inner sep=0] {\includegraphics[width=\w\columnwidth]{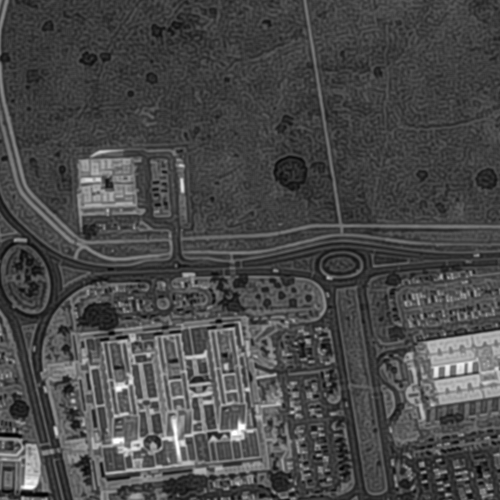}};
    \coordinate (enlarged) at ($(main.south west)+(0,0.5)$);
    \node (enlarged_img) at (enlarged) [anchor=north west,rectangle,draw=line_color,dashed,inner sep=0] {\includegraphics[width=\wenlarged\columnwidth]{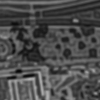}};
    \coordinate (enlarged_size) at ($(enlarged_img.north east)-(enlarged_img.south west)$);
    \coordinate (enlarged_width) at ($(enlarged_img.east)-(enlarged_img.west)$);
    \draw [color=back_line_color,thick] (enlarged_img.north west) rectangle (enlarged_img.south east);
    \draw [color=line_color,thick,dashed] (enlarged_img.north west) rectangle (enlarged_img.south east);

    \coordinate (enlarged_src_BL) at ($(main.north west)+55*(0.126,-0.222)$);
    \draw [color=back_line_color,thick] (enlarged_src_BL) rectangle ($(enlarged_src_BL)+0.21*(enlarged_size)$);
    \draw [color=line_color,dashed,thick] (enlarged_src_BL) rectangle ($(enlarged_src_BL)+0.21*(enlarged_size)$);

    \draw[color=back_line_color,thick,->] ($(enlarged_src_BL)+0.14*(enlarged_width)$) -- (enlarged_img);
    \draw[dashed,color=line_color,thick,->] ($(enlarged_src_BL)+0.14*(enlarged_width)$) -- (enlarged_img);

    \node [below=1 pt of enlarged_img] {\scriptsize EvoIM~\cite{Kawulok2018Gecco}};

\renewcommand{\file}{degraded_fragment2_10_blur_reg_resnet_GASRR.png}
\renewcommand{\zoomfile}{degraded_fragment2_zoom_more_10_blur_reg_resnet_GASRR.png}

    \node (main) at ($(main.north east) + (0.5,0)$) [anchor=north west, inner sep=0] {\includegraphics[width=\w\columnwidth]{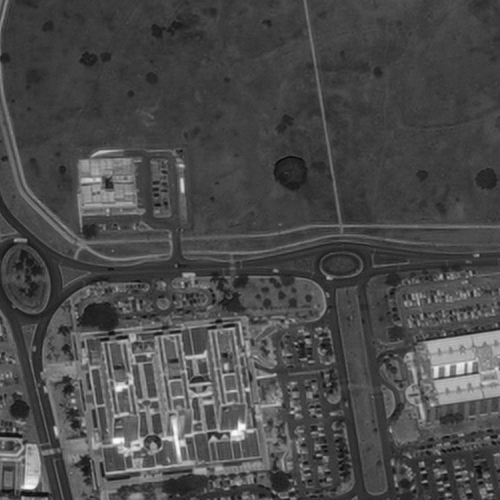}};
    \coordinate (enlarged) at ($(main.south west)+(0,0.5)$);
    \node (enlarged_img) at (enlarged) [anchor=north west,rectangle,draw=line_color,dashed,inner sep=0] {\includegraphics[width=\wenlarged\columnwidth]{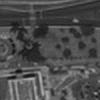}};
    \coordinate (enlarged_size) at ($(enlarged_img.north east)-(enlarged_img.south west)$);
    \coordinate (enlarged_width) at ($(enlarged_img.east)-(enlarged_img.west)$);
    \draw [color=back_line_color,thick] (enlarged_img.north west) rectangle (enlarged_img.south east);
    \draw [color=line_color,thick,dashed] (enlarged_img.north west) rectangle (enlarged_img.south east);

    \coordinate (enlarged_src_BL) at ($(main.north west)+55*(0.126,-0.222)$);
    \draw [color=back_line_color,thick] (enlarged_src_BL) rectangle ($(enlarged_src_BL)+0.21*(enlarged_size)$);
    \draw [color=line_color,dashed,thick] (enlarged_src_BL) rectangle ($(enlarged_src_BL)+0.21*(enlarged_size)$);

    \draw[color=back_line_color,thick,->] ($(enlarged_src_BL)+0.14*(enlarged_width)$) -- (enlarged_img);
    \draw[dashed,color=line_color,thick,->] ($(enlarged_src_BL)+0.14*(enlarged_width)$) -- (enlarged_img);

    \node [below=1 pt of enlarged_img] {\scriptsize \textbf{EvoNet}};

\end{tikzpicture}
}
  \caption{Example of reconstruction from a set of low resolution images (obtained by degrading high resolution images), performed using several SRR methods.}
  \label{fig:degraded_fragment1}
\end{figure*}
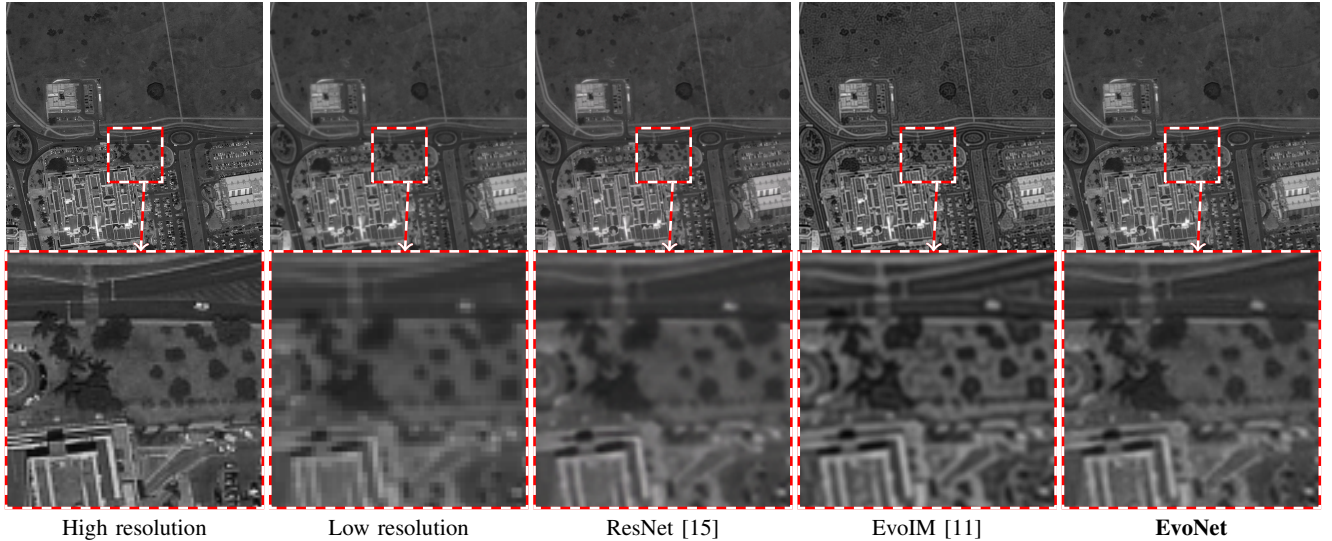

\begin{figure*}[!t]
\renewcommand{\tabcolsep}{0.2cm}
\newcommand{\w}{0.312}
\newcommand{\wenlarged}{0.312}

\newcommand{\file}{real_sat_fragment_rs0010_hr.png}
\newcommand{\zoomfile}{real_sat_fragment_rs0010_zoom_hr.png}

\newcommand{\pos}{0.098,-0.185}
\newcommand{\scale}{0.38}
\resizebox{\textwidth}{!}{
\begin{tikzpicture}[y=.018\columnwidth, x=.018\columnwidth, line/.style={color=blue!70}]
    \colorlet{line_color}{white}
    \colorlet{back_line_color}{red}

    \node (main) at (0,0) [inner sep=0] {\includegraphics[width=\w\columnwidth]{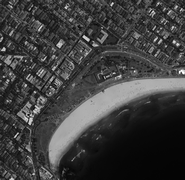}};
    \coordinate (enlarged) at ($(main.south west)+(0,0.5)$);
    \node (enlarged_img) at (enlarged) [anchor=north west,rectangle,draw=line_color,dashed,inner sep=0] {\includegraphics[width=\wenlarged\columnwidth]{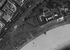}};
    \coordinate (enlarged_size) at ($(enlarged_img.north east)-(enlarged_img.south west)$);
    \coordinate (enlarged_width) at ($(enlarged_img.east)-(enlarged_img.west)$);
    \draw [color=back_line_color,thick] (enlarged_img.north west) rectangle (enlarged_img.south east);
    \draw [color=line_color,thick,dashed] (enlarged_img.north west) rectangle (enlarged_img.south east);

    \coordinate (enlarged_src_BL) at ($(main.north west)+55*(\pos)$);
    \draw [color=back_line_color,thick] (enlarged_src_BL) rectangle ($(enlarged_src_BL)+\scale*(enlarged_size)$);
    \draw [color=line_color,dashed,thick] (enlarged_src_BL) rectangle ($(enlarged_src_BL)+\scale*(enlarged_size)$);

    \draw[color=back_line_color,thick,->] ($(enlarged_src_BL)+0.14*(enlarged_width)$) -- (enlarged_img);
    \draw[dashed,color=line_color,thick,->] ($(enlarged_src_BL)+0.14*(enlarged_width)$) -- (enlarged_img);

    \node [below=1 pt of enlarged_img] {\scriptsize \scriptsize HR};

    \node [left=1 pt of main.north west, anchor=south east, rotate=90] {\scriptsize \emph{Sydney} (Australia)};

    \coordinate (bl) at ($(enlarged_img.south west)$);

\renewcommand{\file}{real_sat_fragment_rs0010_lr.png}
\renewcommand{\zoomfile}{real_sat_fragment_rs0010_zoom_lr.png}

    \node (main) at ($(main.north east) + (0.5,0)$) [anchor=north west, inner sep=0] {\includegraphics[width=\w\columnwidth]{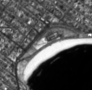}};
    \coordinate (enlarged) at ($(main.south west)+(0,0.5)$);
    \node (enlarged_img) at (enlarged) [anchor=north west,rectangle,draw=line_color,dashed,inner sep=0] {\includegraphics[width=\wenlarged\columnwidth]{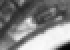}};
    \coordinate (enlarged_size) at ($(enlarged_img.north east)-(enlarged_img.south west)$);
    \coordinate (enlarged_width) at ($(enlarged_img.east)-(enlarged_img.west)$);
    \draw [color=back_line_color,thick] (enlarged_img.north west) rectangle (enlarged_img.south east);
    \draw [color=line_color,thick,dashed] (enlarged_img.north west) rectangle (enlarged_img.south east);

    \coordinate (enlarged_src_BL) at ($(main.north west)+55*(\pos)$);
    \draw [color=back_line_color,thick] (enlarged_src_BL) rectangle ($(enlarged_src_BL)+\scale*(enlarged_size)$);
    \draw [color=line_color,dashed,thick] (enlarged_src_BL) rectangle ($(enlarged_src_BL)+\scale*(enlarged_size)$);

    \draw[color=back_line_color,thick,->] ($(enlarged_src_BL)+0.14*(enlarged_width)$) -- (enlarged_img);
    \draw[dashed,color=line_color,thick,->] ($(enlarged_src_BL)+0.14*(enlarged_width)$) -- (enlarged_img);

    \node [below=1 pt of enlarged_img] {\scriptsize LR};

\renewcommand{\file}{real_sat_fragment_rs0010_resnet.png}
\renewcommand{\zoomfile}{real_sat_fragment_rs0010_zoom_resnet.png}

    \node (main) at ($(main.north east) + (0.5,0)$) [anchor=north west, inner sep=0] {\includegraphics[width=\w\columnwidth]{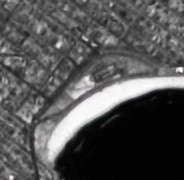}};
    \coordinate (enlarged) at ($(main.south west)+(0,0.5)$);
    \node (enlarged_img) at (enlarged) [anchor=north west,rectangle,draw=line_color,dashed,inner sep=0] {\includegraphics[width=\wenlarged\columnwidth]{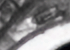}};
    \coordinate (enlarged_size) at ($(enlarged_img.north east)-(enlarged_img.south west)$);
    \coordinate (enlarged_width) at ($(enlarged_img.east)-(enlarged_img.west)$);
    \draw [color=back_line_color,thick] (enlarged_img.north west) rectangle (enlarged_img.south east);
    \draw [color=line_color,thick,dashed] (enlarged_img.north west) rectangle (enlarged_img.south east);

    \coordinate (enlarged_src_BL) at ($(main.north west)+55*(\pos)$);
    \draw [color=back_line_color,thick] (enlarged_src_BL) rectangle ($(enlarged_src_BL)+\scale*(enlarged_size)$);
    \draw [color=line_color,dashed,thick] (enlarged_src_BL) rectangle ($(enlarged_src_BL)+\scale*(enlarged_size)$);

    \draw[color=back_line_color,thick,->] ($(enlarged_src_BL)+0.14*(enlarged_width)$) -- (enlarged_img);
    \draw[dashed,color=line_color,thick,->] ($(enlarged_src_BL)+0.14*(enlarged_width)$) -- (enlarged_img);

    \node [below=1 pt of enlarged_img] {\scriptsize ResNet~\cite{Ledig2017}};

\renewcommand{\file}{real_sat_fragment_rs0010_gasrr_psnr_ls.png}
\renewcommand{\zoomfile}{real_sat_fragment_rs0010_zoom_gasrr_psnr_ls.png}

    \node (main) at ($(main.north east) + (0.5,0)$) [anchor=north west, inner sep=0] {\includegraphics[width=\w\columnwidth]{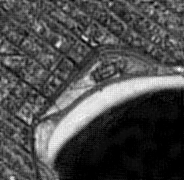}};
    \coordinate (enlarged) at ($(main.south west)+(0,0.5)$);
    \node (enlarged_img) at (enlarged) [anchor=north west,rectangle,draw=line_color,dashed,inner sep=0] {\includegraphics[width=\wenlarged\columnwidth]{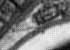}};
    \coordinate (enlarged_size) at ($(enlarged_img.north east)-(enlarged_img.south west)$);
    \coordinate (enlarged_width) at ($(enlarged_img.east)-(enlarged_img.west)$);
    \draw [color=back_line_color,thick] (enlarged_img.north west) rectangle (enlarged_img.south east);
    \draw [color=line_color,thick,dashed] (enlarged_img.north west) rectangle (enlarged_img.south east);

    \coordinate (enlarged_src_BL) at ($(main.north west)+55*(\pos)$);
    \draw [color=back_line_color,thick] (enlarged_src_BL) rectangle ($(enlarged_src_BL)+\scale*(enlarged_size)$);
    \draw [color=line_color,dashed,thick] (enlarged_src_BL) rectangle ($(enlarged_src_BL)+\scale*(enlarged_size)$);

    \draw[color=back_line_color,thick,->] ($(enlarged_src_BL)+0.14*(enlarged_width)$) -- (enlarged_img);
    \draw[dashed,color=line_color,thick,->] ($(enlarged_src_BL)+0.14*(enlarged_width)$) -- (enlarged_img);

    \node [below=1 pt of enlarged_img] {\scriptsize EvoIM~\cite{Kawulok2018Gecco}};

\renewcommand{\file}{real_sat_fragment_rs0010_gasrr_on_resnet_downscaled2x-psnr_hf.png}
\renewcommand{\zoomfile}{real_sat_fragment_rs0010_zoom_gasrr_on_resnet_downscaled2x-psnr_hf.png}

    \node (main) at ($(main.north east) + (0.5,0)$) [anchor=north west, inner sep=0] {\includegraphics[width=\w\columnwidth]{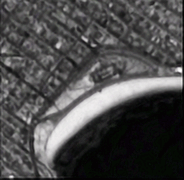}};
    \coordinate (enlarged) at ($(main.south west)+(0,0.5)$);
    \node (enlarged_img) at (enlarged) [anchor=north west,rectangle,draw=line_color,dashed,inner sep=0] {\includegraphics[width=\wenlarged\columnwidth]{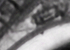}};
    \coordinate (enlarged_size) at ($(enlarged_img.north east)-(enlarged_img.south west)$);
    \coordinate (enlarged_width) at ($(enlarged_img.east)-(enlarged_img.west)$);
    \draw [color=back_line_color,thick] (enlarged_img.north west) rectangle (enlarged_img.south east);
    \draw [color=line_color,thick,dashed] (enlarged_img.north west) rectangle (enlarged_img.south east);

    \coordinate (enlarged_src_BL) at ($(main.north west)+55*(\pos)$);
    \draw [color=back_line_color,thick] (enlarged_src_BL) rectangle ($(enlarged_src_BL)+\scale*(enlarged_size)$);
    \draw [color=line_color,dashed,thick] (enlarged_src_BL) rectangle ($(enlarged_src_BL)+\scale*(enlarged_size)$);

    \draw[color=back_line_color,thick,->] ($(enlarged_src_BL)+0.14*(enlarged_width)$) -- (enlarged_img);
    \draw[dashed,color=line_color,thick,->] ($(enlarged_src_BL)+0.14*(enlarged_width)$) -- (enlarged_img);

    \node [below=1 pt of enlarged_img] {\scriptsize \textbf{EvoNet}};


\renewcommand{\file}{real_sat_fragment_rs0013_hr.png}
\renewcommand{\zoomfile}{real_sat_fragment_rs0013_zoom_hr.png}
\renewcommand{\pos}{0.092,-0.135}
\renewcommand{\scale}{0.44}

    \node (main) at ($(bl.south west) + (0,-1)$) [anchor=north west, inner sep=0] {\includegraphics[width=\w\columnwidth]{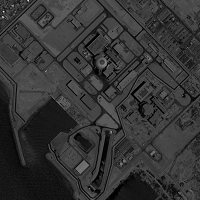}};
    \coordinate (enlarged) at ($(main.south west)+(0,0.5)$);
    \node (enlarged_img) at (enlarged) [anchor=north west,rectangle,draw=line_color,dashed,inner sep=0] {\includegraphics[width=\wenlarged\columnwidth]{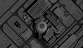}};
    \coordinate (enlarged_size) at ($(enlarged_img.north east)-(enlarged_img.south west)$);
    \coordinate (enlarged_width) at ($(enlarged_img.east)-(enlarged_img.west)$);
    \draw [color=back_line_color,thick] (enlarged_img.north west) rectangle (enlarged_img.south east);
    \draw [color=line_color,thick,dashed] (enlarged_img.north west) rectangle (enlarged_img.south east);

    \coordinate (enlarged_src_BL) at ($(main.north west)+55*(\pos)$);
    \draw [color=back_line_color,thick] (enlarged_src_BL) rectangle ($(enlarged_src_BL)+\scale*(enlarged_size)$);
    \draw [color=line_color,dashed,thick] (enlarged_src_BL) rectangle ($(enlarged_src_BL)+\scale*(enlarged_size)$);

    \draw[color=back_line_color,thick,->] ($(enlarged_src_BL)+0.14*(enlarged_width)$) -- (enlarged_img);
    \draw[dashed,color=line_color,thick,->] ($(enlarged_src_BL)+0.14*(enlarged_width)$) -- (enlarged_img);

    \node [below=1 pt of enlarged_img] {\scriptsize \scriptsize HR};

    \node [left=1 pt of main.north west, anchor=south east, rotate=90] {\scriptsize \emph{Bushehr} (Iran)};

    \coordinate (bl) at ($(enlarged_img.south west)$);

\renewcommand{\file}{real_sat_fragment_rs0013_lr.png}
\renewcommand{\zoomfile}{real_sat_fragment_rs0013_zoom_lr.png}

    \node (main) at ($(main.north east) + (0.5,0)$) [anchor=north west, inner sep=0] {\includegraphics[width=\w\columnwidth]{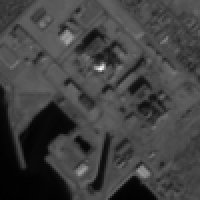}};
    \coordinate (enlarged) at ($(main.south west)+(0,0.5)$);
    \node (enlarged_img) at (enlarged) [anchor=north west,rectangle,draw=line_color,dashed,inner sep=0] {\includegraphics[width=\wenlarged\columnwidth]{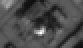}};
    \coordinate (enlarged_size) at ($(enlarged_img.north east)-(enlarged_img.south west)$);
    \coordinate (enlarged_width) at ($(enlarged_img.east)-(enlarged_img.west)$);
    \draw [color=back_line_color,thick] (enlarged_img.north west) rectangle (enlarged_img.south east);
    \draw [color=line_color,thick,dashed] (enlarged_img.north west) rectangle (enlarged_img.south east);

    \coordinate (enlarged_src_BL) at ($(main.north west)+55*(\pos)$);
    \draw [color=back_line_color,thick] (enlarged_src_BL) rectangle ($(enlarged_src_BL)+\scale*(enlarged_size)$);
    \draw [color=line_color,dashed,thick] (enlarged_src_BL) rectangle ($(enlarged_src_BL)+\scale*(enlarged_size)$);

    \draw[color=back_line_color,thick,->] ($(enlarged_src_BL)+0.14*(enlarged_width)$) -- (enlarged_img);
    \draw[dashed,color=line_color,thick,->] ($(enlarged_src_BL)+0.14*(enlarged_width)$) -- (enlarged_img);

    \node [below=1 pt of enlarged_img] {\scriptsize LR};

\renewcommand{\file}{real_sat_fragment_rs0013_resnet.png}
\renewcommand{\zoomfile}{real_sat_fragment_rs0013_zoom_resnet.png}

    \node (main) at ($(main.north east) + (0.5,0)$) [anchor=north west, inner sep=0] {\includegraphics[width=\w\columnwidth]{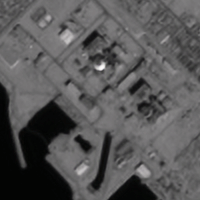}};
    \coordinate (enlarged) at ($(main.south west)+(0,0.5)$);
    \node (enlarged_img) at (enlarged) [anchor=north west,rectangle,draw=line_color,dashed,inner sep=0] {\includegraphics[width=\wenlarged\columnwidth]{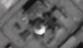}};
    \coordinate (enlarged_size) at ($(enlarged_img.north east)-(enlarged_img.south west)$);
    \coordinate (enlarged_width) at ($(enlarged_img.east)-(enlarged_img.west)$);
    \draw [color=back_line_color,thick] (enlarged_img.north west) rectangle (enlarged_img.south east);
    \draw [color=line_color,thick,dashed] (enlarged_img.north west) rectangle (enlarged_img.south east);

    \coordinate (enlarged_src_BL) at ($(main.north west)+55*(\pos)$);
    \draw [color=back_line_color,thick] (enlarged_src_BL) rectangle ($(enlarged_src_BL)+\scale*(enlarged_size)$);
    \draw [color=line_color,dashed,thick] (enlarged_src_BL) rectangle ($(enlarged_src_BL)+\scale*(enlarged_size)$);

    \draw[color=back_line_color,thick,->] ($(enlarged_src_BL)+0.14*(enlarged_width)$) -- (enlarged_img);
    \draw[dashed,color=line_color,thick,->] ($(enlarged_src_BL)+0.14*(enlarged_width)$) -- (enlarged_img);

    \node [below=1 pt of enlarged_img] {\scriptsize ResNet~\cite{Ledig2017}};

\renewcommand{\file}{real_sat_fragment_rs0013_gasrr_psnr_ls.png}
\renewcommand{\zoomfile}{real_sat_fragment_rs0013_zoom_gasrr_psnr_ls.png}

    \node (main) at ($(main.north east) + (0.5,0)$) [anchor=north west, inner sep=0] {\includegraphics[width=\w\columnwidth]{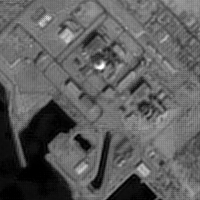}};
    \coordinate (enlarged) at ($(main.south west)+(0,0.5)$);
    \node (enlarged_img) at (enlarged) [anchor=north west,rectangle,draw=line_color,dashed,inner sep=0] {\includegraphics[width=\wenlarged\columnwidth]{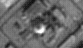}};
    \coordinate (enlarged_size) at ($(enlarged_img.north east)-(enlarged_img.south west)$);
    \coordinate (enlarged_width) at ($(enlarged_img.east)-(enlarged_img.west)$);
    \draw [color=back_line_color,thick] (enlarged_img.north west) rectangle (enlarged_img.south east);
    \draw [color=line_color,thick,dashed] (enlarged_img.north west) rectangle (enlarged_img.south east);

    \coordinate (enlarged_src_BL) at ($(main.north west)+55*(\pos)$);
    \draw [color=back_line_color,thick] (enlarged_src_BL) rectangle ($(enlarged_src_BL)+\scale*(enlarged_size)$);
    \draw [color=line_color,dashed,thick] (enlarged_src_BL) rectangle ($(enlarged_src_BL)+\scale*(enlarged_size)$);

    \draw[color=back_line_color,thick,->] ($(enlarged_src_BL)+0.14*(enlarged_width)$) -- (enlarged_img);
    \draw[dashed,color=line_color,thick,->] ($(enlarged_src_BL)+0.14*(enlarged_width)$) -- (enlarged_img);

    \node [below=1 pt of enlarged_img] {\scriptsize EvoIM~\cite{Kawulok2018Gecco}};

\renewcommand{\file}{real_sat_fragment_rs0013_gasrr_on_resnet_downscaled2x-psnr_hf.png}
\renewcommand{\zoomfile}{real_sat_fragment_rs0013_zoom_gasrr_on_resnet_downscaled2x-psnr_hf.png}

    \node (main) at ($(main.north east) + (0.5,0)$) [anchor=north west, inner sep=0] {\includegraphics[width=\w\columnwidth]{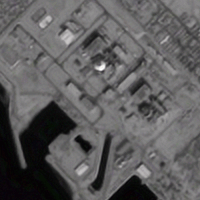}};
    \coordinate (enlarged) at ($(main.south west)+(0,0.5)$);
    \node (enlarged_img) at (enlarged) [anchor=north west,rectangle,draw=line_color,dashed,inner sep=0] {\includegraphics[width=\wenlarged\columnwidth]{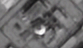}};
    \coordinate (enlarged_size) at ($(enlarged_img.north east)-(enlarged_img.south west)$);
    \coordinate (enlarged_width) at ($(enlarged_img.east)-(enlarged_img.west)$);
    \draw [color=back_line_color,thick] (enlarged_img.north west) rectangle (enlarged_img.south east);
    \draw [color=line_color,thick,dashed] (enlarged_img.north west) rectangle (enlarged_img.south east);

    \coordinate (enlarged_src_BL) at ($(main.north west)+55*(\pos)$);
    \draw [color=back_line_color,thick] (enlarged_src_BL) rectangle ($(enlarged_src_BL)+\scale*(enlarged_size)$);
    \draw [color=line_color,dashed,thick] (enlarged_src_BL) rectangle ($(enlarged_src_BL)+\scale*(enlarged_size)$);

    \draw[color=back_line_color,thick,->] ($(enlarged_src_BL)+0.14*(enlarged_width)$) -- (enlarged_img);
    \draw[dashed,color=line_color,thick,->] ($(enlarged_src_BL)+0.14*(enlarged_width)$) -- (enlarged_img);

    \node [below=1 pt of enlarged_img] {\scriptsize \textbf{EvoNet}};


\renewcommand{\file}{real_sat_fragment_rs0014_hr.png}
\renewcommand{\zoomfile}{real_sat_fragment_rs0014_zoom_hr.png}
\renewcommand{\pos}{0.02,-0.141}
\renewcommand{\scale}{0.43}

    \node (main) at ($(bl.south west) + (0,-1)$) [anchor=north west, inner sep=0] {\includegraphics[width=\w\columnwidth]{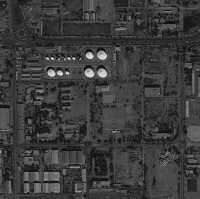}};
    \coordinate (enlarged) at ($(main.south west)+(0,0.5)$);
    \node (enlarged_img) at (enlarged) [anchor=north west,rectangle,draw=line_color,dashed,inner sep=0] {\includegraphics[width=\wenlarged\columnwidth]{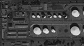}};
    \coordinate (enlarged_size) at ($(enlarged_img.north east)-(enlarged_img.south west)$);
    \coordinate (enlarged_width) at ($(enlarged_img.east)-(enlarged_img.west)$);
    \draw [color=back_line_color,thick] (enlarged_img.north west) rectangle (enlarged_img.south east);
    \draw [color=line_color,thick,dashed] (enlarged_img.north west) rectangle (enlarged_img.south east);

    \coordinate (enlarged_src_BL) at ($(main.north west)+55*(\pos)$);
    \draw [color=back_line_color,thick] (enlarged_src_BL) rectangle ($(enlarged_src_BL)+\scale*(enlarged_size)$);
    \draw [color=line_color,dashed,thick] (enlarged_src_BL) rectangle ($(enlarged_src_BL)+\scale*(enlarged_size)$);

    \draw[color=back_line_color,thick,->] ($(enlarged_src_BL)+0.14*(enlarged_width)$) -- (enlarged_img);
    \draw[dashed,color=line_color,thick,->] ($(enlarged_src_BL)+0.14*(enlarged_width)$) -- (enlarged_img);

    \node [below=1 pt of enlarged_img] {\scriptsize \scriptsize High resolution};

    \node [left=1 pt of main.north west, anchor=south east, rotate=90] {\scriptsize \emph{Bandar Abbas} (Iran)};

    \coordinate (bl) at ($(enlarged_img.south west)$);

\renewcommand{\file}{real_sat_fragment_rs0014_lr.png}
\renewcommand{\zoomfile}{real_sat_fragment_rs0014_zoom_lr.png}

    \node (main) at ($(main.north east) + (0.5,0)$) [anchor=north west, inner sep=0] {\includegraphics[width=\w\columnwidth]{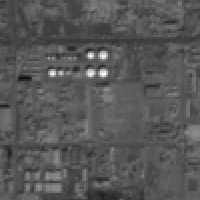}};
    \coordinate (enlarged) at ($(main.south west)+(0,0.5)$);
    \node (enlarged_img) at (enlarged) [anchor=north west,rectangle,draw=line_color,dashed,inner sep=0] {\includegraphics[width=\wenlarged\columnwidth]{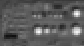}};
    \coordinate (enlarged_size) at ($(enlarged_img.north east)-(enlarged_img.south west)$);
    \coordinate (enlarged_width) at ($(enlarged_img.east)-(enlarged_img.west)$);
    \draw [color=back_line_color,thick] (enlarged_img.north west) rectangle (enlarged_img.south east);
    \draw [color=line_color,thick,dashed] (enlarged_img.north west) rectangle (enlarged_img.south east);

    \coordinate (enlarged_src_BL) at ($(main.north west)+55*(\pos)$);
    \draw [color=back_line_color,thick] (enlarged_src_BL) rectangle ($(enlarged_src_BL)+\scale*(enlarged_size)$);
    \draw [color=line_color,dashed,thick] (enlarged_src_BL) rectangle ($(enlarged_src_BL)+\scale*(enlarged_size)$);

    \draw[color=back_line_color,thick,->] ($(enlarged_src_BL)+0.14*(enlarged_width)$) -- (enlarged_img);
    \draw[dashed,color=line_color,thick,->] ($(enlarged_src_BL)+0.14*(enlarged_width)$) -- (enlarged_img);

    \node [below=1 pt of enlarged_img] {\scriptsize Low resolution};

\renewcommand{\file}{real_sat_fragment_rs0014_resnet.png}
\renewcommand{\zoomfile}{real_sat_fragment_rs0014_zoom_resnet.png}

    \node (main) at ($(main.north east) + (0.5,0)$) [anchor=north west, inner sep=0] {\includegraphics[width=\w\columnwidth]{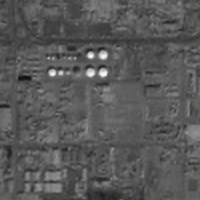}};
    \coordinate (enlarged) at ($(main.south west)+(0,0.5)$);
    \node (enlarged_img) at (enlarged) [anchor=north west,rectangle,draw=line_color,dashed,inner sep=0] {\includegraphics[width=\wenlarged\columnwidth]{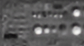}};
    \coordinate (enlarged_size) at ($(enlarged_img.north east)-(enlarged_img.south west)$);
    \coordinate (enlarged_width) at ($(enlarged_img.east)-(enlarged_img.west)$);
    \draw [color=back_line_color,thick] (enlarged_img.north west) rectangle (enlarged_img.south east);
    \draw [color=line_color,thick,dashed] (enlarged_img.north west) rectangle (enlarged_img.south east);

    \coordinate (enlarged_src_BL) at ($(main.north west)+55*(\pos)$);
    \draw [color=back_line_color,thick] (enlarged_src_BL) rectangle ($(enlarged_src_BL)+\scale*(enlarged_size)$);
    \draw [color=line_color,dashed,thick] (enlarged_src_BL) rectangle ($(enlarged_src_BL)+\scale*(enlarged_size)$);

    \draw[color=back_line_color,thick,->] ($(enlarged_src_BL)+0.14*(enlarged_width)$) -- (enlarged_img);
    \draw[dashed,color=line_color,thick,->] ($(enlarged_src_BL)+0.14*(enlarged_width)$) -- (enlarged_img);

    \node [below=1 pt of enlarged_img] {\scriptsize ResNet~\cite{Ledig2017}};

\renewcommand{\file}{real_sat_fragment_rs0014_gasrr_psnr_ls.png}
\renewcommand{\zoomfile}{real_sat_fragment_rs0014_zoom_gasrr_psnr_ls.png}

    \node (main) at ($(main.north east) + (0.5,0)$) [anchor=north west, inner sep=0] {\includegraphics[width=\w\columnwidth]{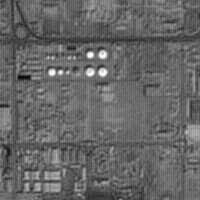}};
    \coordinate (enlarged) at ($(main.south west)+(0,0.5)$);
    \node (enlarged_img) at (enlarged) [anchor=north west,rectangle,draw=line_color,dashed,inner sep=0] {\includegraphics[width=\wenlarged\columnwidth]{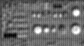}};
    \coordinate (enlarged_size) at ($(enlarged_img.north east)-(enlarged_img.south west)$);
    \coordinate (enlarged_width) at ($(enlarged_img.east)-(enlarged_img.west)$);
    \draw [color=back_line_color,thick] (enlarged_img.north west) rectangle (enlarged_img.south east);
    \draw [color=line_color,thick,dashed] (enlarged_img.north west) rectangle (enlarged_img.south east);

    \coordinate (enlarged_src_BL) at ($(main.north west)+55*(\pos)$);
    \draw [color=back_line_color,thick] (enlarged_src_BL) rectangle ($(enlarged_src_BL)+\scale*(enlarged_size)$);
    \draw [color=line_color,dashed,thick] (enlarged_src_BL) rectangle ($(enlarged_src_BL)+\scale*(enlarged_size)$);

    \draw[color=back_line_color,thick,->] ($(enlarged_src_BL)+0.14*(enlarged_width)$) -- (enlarged_img);
    \draw[dashed,color=line_color,thick,->] ($(enlarged_src_BL)+0.14*(enlarged_width)$) -- (enlarged_img);

    \node [below=1 pt of enlarged_img] {\scriptsize EvoIM~\cite{Kawulok2018Gecco}};

\renewcommand{\file}{real_sat_fragment_rs0014_gasrr_on_resnet_downscaled2x-psnr_hf.png}
\renewcommand{\zoomfile}{real_sat_fragment_rs0014_zoom_gasrr_on_resnet_downscaled2x-psnr_hf.png}

    \node (main) at ($(main.north east) + (0.5,0)$) [anchor=north west, inner sep=0] {\includegraphics[width=\w\columnwidth]{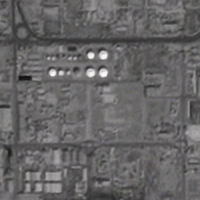}};
    \coordinate (enlarged) at ($(main.south west)+(0,0.5)$);
    \node (enlarged_img) at (enlarged) [anchor=north west,rectangle,draw=line_color,dashed,inner sep=0] {\includegraphics[width=\wenlarged\columnwidth]{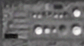}};
    \coordinate (enlarged_size) at ($(enlarged_img.north east)-(enlarged_img.south west)$);
    \coordinate (enlarged_width) at ($(enlarged_img.east)-(enlarged_img.west)$);
    \draw [color=back_line_color,thick] (enlarged_img.north west) rectangle (enlarged_img.south east);
    \draw [color=line_color,thick,dashed] (enlarged_img.north west) rectangle (enlarged_img.south east);

    \coordinate (enlarged_src_BL) at ($(main.north west)+55*(\pos)$);
    \draw [color=back_line_color,thick] (enlarged_src_BL) rectangle ($(enlarged_src_BL)+\scale*(enlarged_size)$);
    \draw [color=line_color,dashed,thick] (enlarged_src_BL) rectangle ($(enlarged_src_BL)+\scale*(enlarged_size)$);

    \draw[color=back_line_color,thick,->] ($(enlarged_src_BL)+0.14*(enlarged_width)$) -- (enlarged_img);
    \draw[dashed,color=line_color,thick,->] ($(enlarged_src_BL)+0.14*(enlarged_width)$) -- (enlarged_img);

    \node [below=1 pt of enlarged_img] {\scriptsize \textbf{EvoNet}};

\end{tikzpicture}

}
  \caption{Reconstruction of real satellite images, performed using different SRR methods (high resolution images are given for reference). }
  \label{fig:real_sat}
\end{figure*}
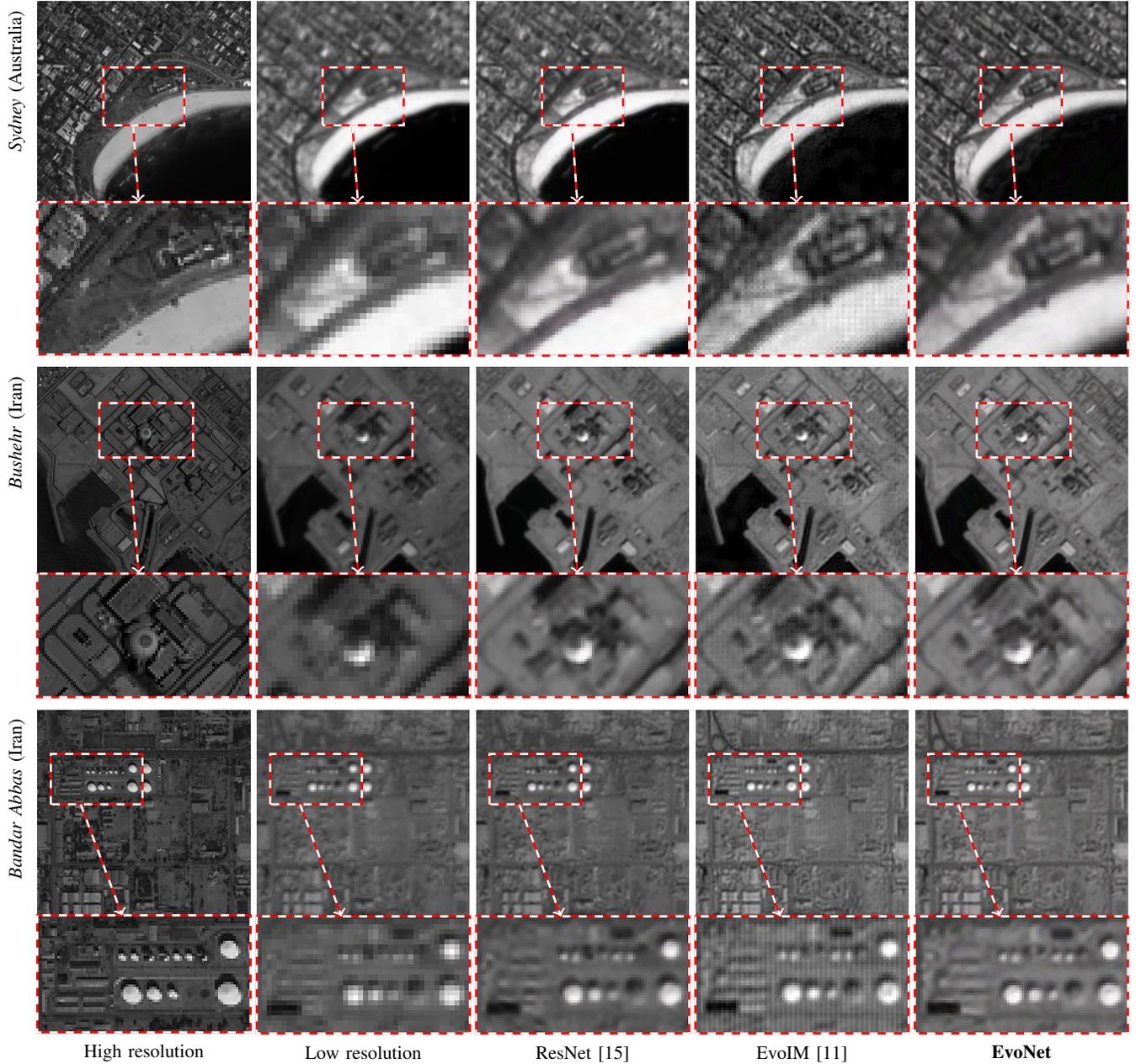

\begin{figure}[!b]
\renewcommand{\tabcolsep}{0.2cm}
\newcommand{\w}{0.312}
\newcommand{\wenlarged}{0.312}

\newcommand{\file}{moon_crop_gasrr_on_resnet_4x.png}
\newcommand{\zoomfile}{moon_zoom_gasrr_on_resnet_4x.png}

\newcommand{\pos}{0.14,-0.055}
\newcommand{\scale}{0.31}
\resizebox{1.02\columnwidth}{!}{
\begin{tikzpicture}[y=.018\columnwidth, x=.018\columnwidth, line/.style={color=blue!70}]
    \colorlet{line_color}{white}
    \colorlet{back_line_color}{red}

    \node (main) at (0,0) [inner sep=0] {\includegraphics[width=\w\columnwidth]{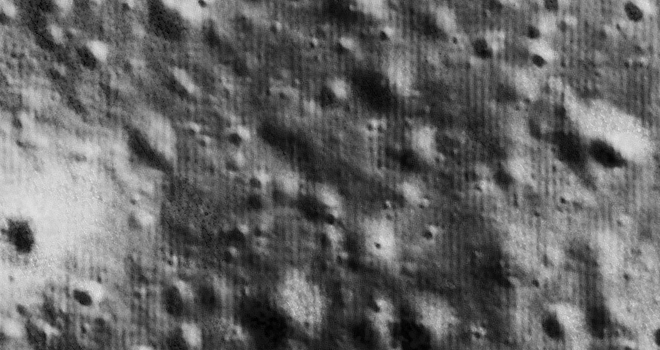}};
    \coordinate (enlarged) at ($(main.south west)+(0,0.5)$);
    \node (enlarged_img) at (enlarged) [anchor=north west,rectangle,draw=line_color,dashed,inner sep=0] {\includegraphics[width=\wenlarged\columnwidth]{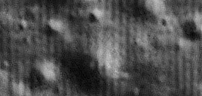}};
    \coordinate (enlarged_size) at ($(enlarged_img.north east)-(enlarged_img.south west)$);
    \coordinate (enlarged_width) at ($(enlarged_img.east)-(enlarged_img.west)$);
    \draw [color=back_line_color,thick] (enlarged_img.north west) rectangle (enlarged_img.south east);
    \draw [color=line_color,thick,dashed] (enlarged_img.north west) rectangle (enlarged_img.south east);

    \coordinate (enlarged_src_BL) at ($(main.north west)+55*(\pos)$);
    \draw [color=back_line_color,thick] (enlarged_src_BL) rectangle ($(enlarged_src_BL)+\scale*(enlarged_size)$);
    \draw [color=line_color,dashed,thick] (enlarged_src_BL) rectangle ($(enlarged_src_BL)+\scale*(enlarged_size)$);

    \draw[color=back_line_color,thick,->] ($(enlarged_src_BL)+0.14*(enlarged_width)$) -- (enlarged_img);
    \draw[dashed,color=line_color,thick,->] ($(enlarged_src_BL)+0.14*(enlarged_width)$) -- (enlarged_img);


    \coordinate (bl) at ($(enlarged_img.south west)$);

\renewcommand{\file}{moon_crop_resnet4x.png}
\renewcommand{\zoomfile}{moon_zoom_resnet4x.png}

    \node (main) at ($(main.north east) + (0.5,0)$) [anchor=north west, inner sep=0] {\includegraphics[width=\w\columnwidth]{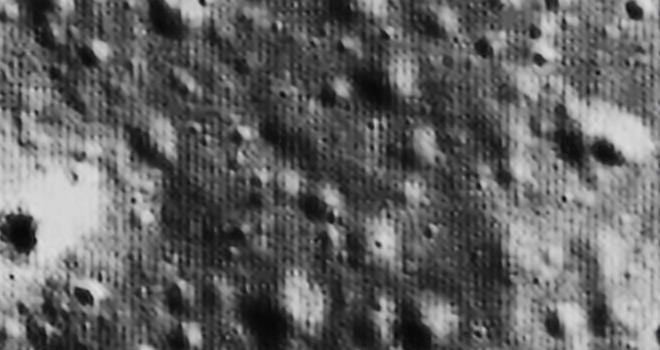}};
    \coordinate (enlarged) at ($(main.south west)+(0,0.5)$);
    \node (enlarged_img) at (enlarged) [anchor=north west,rectangle,draw=line_color,dashed,inner sep=0] {\includegraphics[width=\wenlarged\columnwidth]{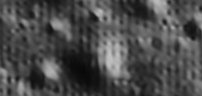}};
    \coordinate (enlarged_size) at ($(enlarged_img.north east)-(enlarged_img.south west)$);
    \coordinate (enlarged_width) at ($(enlarged_img.east)-(enlarged_img.west)$);
    \draw [color=back_line_color,thick] (enlarged_img.north west) rectangle (enlarged_img.south east);
    \draw [color=line_color,thick,dashed] (enlarged_img.north west) rectangle (enlarged_img.south east);

    \coordinate (enlarged_src_BL) at ($(main.north west)+55*(\pos)$);
    \draw [color=back_line_color,thick] (enlarged_src_BL) rectangle ($(enlarged_src_BL)+\scale*(enlarged_size)$);
    \draw [color=line_color,dashed,thick] (enlarged_src_BL) rectangle ($(enlarged_src_BL)+\scale*(enlarged_size)$);

    \draw[color=back_line_color,thick,->] ($(enlarged_src_BL)+0.14*(enlarged_width)$) -- (enlarged_img);
    \draw[dashed,color=line_color,thick,->] ($(enlarged_src_BL)+0.14*(enlarged_width)$) -- (enlarged_img);


\renewcommand{\file}{moon_crop_gasrr.png}
\renewcommand{\zoomfile}{moon_zoom_gasrr.png}

    \node (main) at ($(bl.south west) + (0,-2)$) [anchor=north west, inner sep=0] {\includegraphics[width=\w\columnwidth]{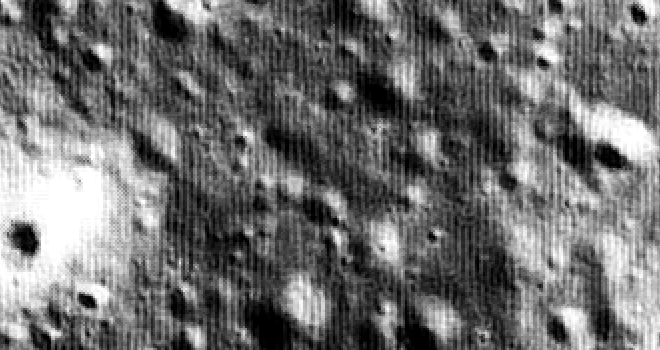}};
    \coordinate (enlarged) at ($(main.south west)+(0,0.5)$);
    \node (enlarged_img) at (enlarged) [anchor=north west,rectangle,draw=line_color,dashed,inner sep=0] {\includegraphics[width=\wenlarged\columnwidth]{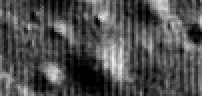}};
    \coordinate (enlarged_size) at ($(enlarged_img.north east)-(enlarged_img.south west)$);
    \coordinate (enlarged_width) at ($(enlarged_img.east)-(enlarged_img.west)$);
    \draw [color=back_line_color,thick] (enlarged_img.north west) rectangle (enlarged_img.south east);
    \draw [color=line_color,thick,dashed] (enlarged_img.north west) rectangle (enlarged_img.south east);

    \coordinate (enlarged_src_BL) at ($(main.north west)+55*(\pos)$);
    \draw [color=back_line_color,thick] (enlarged_src_BL) rectangle ($(enlarged_src_BL)+\scale*(enlarged_size)$);
    \draw [color=line_color,dashed,thick] (enlarged_src_BL) rectangle ($(enlarged_src_BL)+\scale*(enlarged_size)$);

    \draw[color=back_line_color,thick,->] ($(enlarged_src_BL)+0.14*(enlarged_width)$) -- (enlarged_img);
    \draw[dashed,color=line_color,thick,->] ($(enlarged_src_BL)+0.14*(enlarged_width)$) -- (enlarged_img);


\renewcommand{\file}{moon_crop_lr.png}
\renewcommand{\zoomfile}{moon_zoom_lr.png}

    \node (main) at ($(main.north east) + (0.5,0)$) [anchor=north west, inner sep=0] {\includegraphics[width=\w\columnwidth]{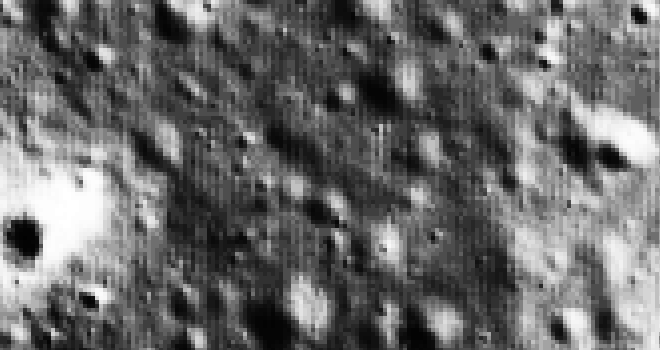}};
    \coordinate (enlarged) at ($(main.south west)+(0,0.5)$);
    \node (enlarged_img) at (enlarged) [anchor=north west,rectangle,draw=line_color,dashed,inner sep=0] {\includegraphics[width=\wenlarged\columnwidth]{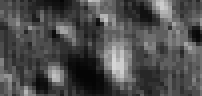}};
    \coordinate (enlarged_size) at ($(enlarged_img.north east)-(enlarged_img.south west)$);
    \coordinate (enlarged_width) at ($(enlarged_img.east)-(enlarged_img.west)$);
    \draw [color=back_line_color,thick] (enlarged_img.north west) rectangle (enlarged_img.south east);
    \draw [color=line_color,thick,dashed] (enlarged_img.north west) rectangle (enlarged_img.south east);

    \coordinate (enlarged_src_BL) at ($(main.north west)+55*(\pos)$);
    \draw [color=back_line_color,thick] (enlarged_src_BL) rectangle ($(enlarged_src_BL)+\scale*(enlarged_size)$);
    \draw [color=line_color,dashed,thick] (enlarged_src_BL) rectangle ($(enlarged_src_BL)+\scale*(enlarged_size)$);

    \draw[color=back_line_color,thick,->] ($(enlarged_src_BL)+0.14*(enlarged_width)$) -- (enlarged_img);
    \draw[dashed,color=line_color,thick,->] ($(enlarged_src_BL)+0.14*(enlarged_width)$) -- (enlarged_img);


\end{tikzpicture}

}
   \caption{Reconstruction from Lunar Reconnaissance Orbiter Camera images ($N=6$ LR observations) performed using different SRR methods.}
   \label{fig:moon}

\end{figure}

\section{Experiments} \label{sec:exp}

For validation, we used three types of data in the test set, namely: (i)~\emph{artificially-degraded} (AD) images---10 scenes, for each a set $\lrset$ obtained from an HR image $\hr$ with $N=4$ different subpixel shifts applied before further degradation, each $\lri$ of size $500\times 500$ pixels, (ii)~\emph{real satellite} (RS$+$) images of the same region, acquired at different resolution---we used three Sentinel-2 scenes as LR ($N=10$ LR images in each scene), two of which are matched with SPOT images (presenting \emph{Bushehr}, Iran, LR of size $300\times 291$ pixels, and \emph{Bandar Abbas}, Iran, $240\times 266$ pixels) and one is matched with Digital Globe WorldView-4 image (\emph{Sydney}, Australia, $92\times 90$ pixels), and (iii)~real satellite images available without any higher-resolution reference (RS$-$, over 20 scenes). For AD and RS$+$, we quantify the reconstruction quality based on the similarity between $\hr$ and $\hrsrr$, and for RS$-$, we rely exclusively on subjective qualitative assessment (as no reference is available). The reconstruction outcome is evaluated quantitatively at the dimensions $2\times$ larger than for input LR images (EvoNet and ResNet enlarge LR images $4\times$, so  we downscale these outcomes $2\times$ for fair comparison with the remaining methods). For RS$+$, $\hrsrr$ is compared with Digital Globe and SPOT images, downscaled to fit the dimensions of $\hrsrr$. In addition to \emph{peak signal-to-noise ratio} (PSNR) and \emph{structural similarity index} (SSIM), we measure the similarity using more advanced metrics~\cite{Benecki2018AA}: \emph{information fidelity criterion} (IFC), \emph{visual information fidelity} (VIF), \emph{universal image quality index} (UIQI), and PSNR for images treated with a high-pass filter (PSNR\textsubscript{HF}) and local standard deviation (PSNR\textsubscript{LS}). For all these metrics, higher values indicate higher similarity between the reconstruction outcome and the reference image.

EvoNet is compared with two single-image SRR methods: SRR based on wavelet transform (SR-DWT)~\cite{Demirel2011} and ResNet~\cite{Ledig2017}, and with three multiple-image ones: GPA~\cite{Schultz1996}, SR-ADE~\cite{Zhu2016}, and EvoIM~\cite{Kawulok2018Gecco}. EvoIM (also exploited in EvoNet) was trained separately for artificially-degraded images and for real satellite data, as reported in~\cite{Kawulok2018Gecco}, using PSNR\textsubscript{HF}~\cite{Benecki2018AA} as the fitness function (there were no overlaps between training and test sets). ResNet was trained using images from the DIV2K dataset\footnote{DIV2K dataset is available at \url{https://data.vision.ee.ethz.ch/cvl/DIV2K}}. We implemented all the investigated algorithms in C++, and we used Python with Keras to implement ResNet. The experiments were run on an Intel i5 3.2 GHz computer with 16 GB RAM, and ResNet was trained on a GTX 1060 6 GB GPU.

\begin{table}[!h]
\centering
\caption{Reconstruction accuracy and processing times for artificially degraded images (best scores are marked as bold).}\label{tab:measures_degraded}
\scriptsize
\renewcommand{\tabcolsep}{1.1mm}
\begin{tabular}{rcccccccc}
\Xhline{2\arrayrulewidth}
Algorithm &
  IFC & PSNR & PSNR\textsubscript{HF} & PSNR\textsubscript{LS} & SSIM & UIQI & VIF & Time (s)\\
\hline
SR-DWT~\cite{Demirel2011}          & 2.281 & 28.833 & 40.580 & 38.613 & 0.813 & 0.757 & 0.458 & 4 \\
ResNet~\cite{Ledig2017}         & 2.517 & 28.773 & 34.038 & 33.470 & 0.823 & 0.749 & 0.453 & 30  \\
GPA~\cite{Schultz1996}          & 2.436 & 28.054 & 32.924 & 32.522 & 0.792 & 0.712 & 0.422 & 15 \\
SR-ADE~\cite{Zhu2016}           & 2.289 & 27.237 & 32.049 & 31.742 & 0.756 & 0.666 & 0.378 & 17 \\
EvoIM~\cite{Kawulok2018Gecco}   & 3.190 & 31.185 & 39.067 & 38.166 & 0.863 & 0.801 & 0.561 & 4 \\
\algname$^{\rm A}$              & 2.979 & 32.929 & 41.522 & 41.437 & 0.919 & 0.864 & 0.596 & 161 \\
\algname                        & \textbf{3.256} & \textbf{35.065} & \textbf{44.839} & \textbf{44.645} & \textbf{0.948} & \textbf{0.902} & \textbf{0.661} & 118 \\
\Xhline{2\arrayrulewidth}
\multicolumn{8}{l}{\algname$^{\rm A}$---image registration performed for ResNet outputs}\\
\end{tabular}
\end{table}
In Table~\ref{tab:measures_degraded}, we report the reconstruction accuracy for AD images alongside the processing times. For fair comparison, all the reconstruction tests were run on a CPU, which explains long times of ResNet and EvoNet (GPU was used only for training ResNet). EvoNet allows for the most accurate reconstruction, rendering consistently best scores, and multiple-image EvoIM renders higher scores than single-image SR-DWT and ResNet. Examples of reconstruction are presented in Fig.~\ref{fig:degraded_fragment1}---the outcome of ResNet is more blurred than EvoNet, with less details visible, and EvoIM produces definitely more artifacts; overall, EvoNet renders very plausible outcome, which most resembles the HR image. We have also tried to register the images after they are processed with ResNet---as expected, this decreases the reconstruction accuracy, while extending the processing time (see Table~\ref{tab:measures_degraded}).

Quantitative results obtained for RS$+$ images are reported in Table~\ref{tab:measures_real} (we also show the values averaged over three images). It can be seen that for \emph{Sydney} and \emph{Bandar Abbas}, EvoNet renders highest scores for most metrics (including IFC and VIF which were found most meaningful for assessing SRR~\cite{Benecki2018AA}). For \emph{Bushehr}, the scores differ less among the methods, and the metrics are not consistent in indicating the most accurate method---possibly because this image contains more plain areas compared with two remaining scenes. Average PSNR is highest for ResNet, which can be caused by using MSE as the loss function for training (PSNR is based on MSE). All other metrics indicate that EvoNet outperforms the remaining methods. From Fig.~\ref{fig:real_sat}, it can be seen that the quantitative results are coherent with the visual assessment---all the methods increase the interpretation capacities compared with LR, and the outcome obtained using EvoNet recovers more details than ResNet, without introducing the artifacts visible for EvoIM.

The outcomes obtained for RS$-$ images (without any HR reference) generally confirm our observations discussed for RS$+$ images. In Fig.~\ref{fig:moon}, we show an interesting example of reconstruction from Lunar Reconnaissance Orbiter Camera images. It is worth noting that these LR images contain some artifacts in a form of faint vertical stripes, which result from the sensor characteristics (the images were not preprocessed). In this case, not only does EvoNet render the highest reconstruction quality, but it also manages to make these artifacts less apparent compared with EvoIM and ResNet (this can be explained by the fact that ResNet changes the artifacts to be grid-like, which can be further reduced during the fusion).

\section{Conclusions} \label{sec:concl}

In this letter, we proposed a novel method for multiple-image super-resolution which exploits the recent advancements in deep learning. We demonstrated that the ResNet deep CNN applied to enhance each individual LR image before performing the multiple-image fusion, can substantially improve the final super-resolved image. The reported quantitative and qualitative results indicate that the proposed approach is highly competitive with the state of the art both in single-image SRR, as well as in multiple-image super-resolution.

Our ongoing work is aimed at developing deep architectures for learning the entire process of multiple-image reconstruction, possibly including image registration. 


\begin{table}[!h]
\centering
\caption{Reconstruction accuracy for three Sentinel-2 images (the best scores are marked as bold).}\label{tab:measures_real}
\scriptsize
\renewcommand{\tabcolsep}{1.3mm}

\begin{tabular}{rrccccccc}
\Xhline{2\arrayrulewidth}
& Algorithm &
 IFC & PSNR & PSNR\textsubscript{HF} & PSNR\textsubscript{LS} & SSIM & UIQI & VIF \\
\hline
\multirow{6}{*}{\begin{sideways}\emph{Sydney}\end{sideways}} &
SR-DWT~\cite{Demirel2011}              & 1.146 & 14.883 & 32.306 & 29.717 & 0.345 & 0.284 & 0.125 \\
& ResNet~\cite{Ledig2017}           & 1.070 & 14.533 & 32.609 & 31.029 & 0.292 & 0.176 & 0.105 \\
& GPA~\cite{Schultz1996}            & 1.191 & 16.619 & 31.928 & 30.710 & 0.398 & 0.236 & 0.121 \\
& SR-ADE~\cite{Zhu2016}             & 1.375 & \textbf{17.250} & 30.349 & 29.289 & 0.467 & 0.302 & 0.132 \\
& EvoIM~\cite{Kawulok2018Gecco}     & 1.271 & 16.384 & \textbf{34.657} & 32.560 & 0.429 & 0.314 & 0.129 \\
& \algname                          & \textbf{1.387} & 16.722 & 34.349 & \textbf{32.607} & \textbf{0.487} & \textbf{0.334} & \textbf{0.139} \\
\hline
\multirow{6}{*}{\begin{sideways}\emph{Bushehr}\end{sideways}}
& SR-DWT~\cite{Demirel2011}            & 1.032 & 15.432 & 36.475 & 34.403 & 0.344 & 0.199 & 0.087 \\
& ResNet~\cite{Ledig2017}           & 1.194 & \textbf{15.481} & 37.072 & \textbf{35.997} & 0.424 & 0.233 & 0.098 \\
& GPA~\cite{Schultz1996}            & \textbf{1.285} & 14.827 & 35.135 & 34.168 & \textbf{0.474} & 0.253 & \textbf{0.114} \\
& SR-ADE~\cite{Zhu2016}             & 1.185 & 14.704 & 33.804 & 32.963 & 0.458 & 0.218 & 0.102 \\
&     EvoIM~\cite{Kawulok2018Gecco} & 1.134 & 14.470 & \textbf{37.237} & 35.956 & 0.362 & 0.227 & 0.098 \\
&     \algname                      & 1.261 & 14.528 & 36.878 & 35.739 & 0.433 & \textbf{0.261} & 0.109 \\
\hline
\multirow{6}{*}{\begin{sideways}\emph{Bandar Abbas}\end{sideways}}
& SR-DWT~\cite{Demirel2011}            & 1.031 & 18.697 & 36.021 & 34.542 & 0.419 & 0.221 & 0.092 \\
& ResNet~\cite{Ledig2017}           & 1.395 & \textbf{19.385} & 38.714 & 37.657 & \textbf{0.561} & 0.292 & 0.130 \\
& GPA~\cite{Schultz1996}            & 1.419 & 16.414 & 35.736 & 34.900 & 0.551 & 0.292 & 0.140 \\
& SR-ADE~\cite{Zhu2016}             & 1.305 & 16.187 & 33.381 & 32.634 & 0.521 & 0.249 & 0.124 \\
&     EvoIM~\cite{Kawulok2018Gecco} & 1.148 & 16.068 & 37.158 & 35.909 & 0.414 & 0.255 & 0.114 \\
&     \algname                      & \textbf{1.494} & 16.226 & \textbf{39.350} & \textbf{38.162} & 0.527 & \textbf{0.318} & \textbf{0.153} \\
\hline
\multirow{6}{*}{\begin{sideways}Mean scores\end{sideways}}
& SR-DWT~\cite{Demirel2011}            & 1.070 & 16.337 & 34.934 & 32.887 & 0.369 & 0.234 & 0.101 \\
& ResNet~\cite{Ledig2017}           & 1.220 & \textbf{16.467} & 36.132 & 34.894 & 0.426 & 0.234 & 0.111 \\
& GPA~\cite{Schultz1996}            & 1.299 & 15.953 & 34.266 & 33.259 & 0.474 & 0.260 & 0.125 \\
& SR-ADE~\cite{Zhu2016}             & 1.288 & 16.047 & 32.512 & 31.629 & \textbf{0.482} & 0.256 & 0.119 \\
& EvoIM~\cite{Kawulok2018Gecco}     & 1.184 & 15.641 & 36.351 & 34.809 & 0.402 & 0.265 & 0.114 \\
& \algname                          & \textbf{1.381} & 15.825 & \textbf{36.859} & \textbf{35.503} & \textbf{0.482} & \textbf{0.304} & \textbf{0.134} \\
\Xhline{2\arrayrulewidth}
\end{tabular}

\end{table}

%


\begin{thebibliography}{10}
\providecommand{\url}[1]{#1}
\csname url@samestyle\endcsname
\providecommand{\newblock}{\relax}
\providecommand{\bibinfo}[2]{#2}
\providecommand{\BIBentrySTDinterwordspacing}{\spaceskip=0pt\relax}
\providecommand{\BIBentryALTinterwordstretchfactor}{4}
\providecommand{\BIBentryALTinterwordspacing}{\spaceskip=\fontdimen2\font plus
\BIBentryALTinterwordstretchfactor\fontdimen3\font minus
  \fontdimen4\font\relax}
\providecommand{\BIBforeignlanguage}[2]{{%
\expandafter\ifx\csname l@#1\endcsname\relax
\typeout{** WARNING: IEEEtranS.bst: No hyphenation pattern has been}%
\typeout{** loaded for the language `#1'. Using the pattern for}%
\typeout{** the default language instead.}%
\else
\language=\csname l@#1\endcsname
\fi
#2}}
\providecommand{\BIBdecl}{\relax}
\BIBdecl

\bibitem{Akgun2005}
T.~Akgun, Y.~Altunbasak, and R.~M. Mersereau, ``Super-resolution reconstruction
  of hyperspectral images,'' \emph{IEEE Trans. on Image Process.}, vol.~14,
  no.~11, pp. 1860--1875, 2005.

\bibitem{Benecki2018AA}
P.~Benecki, M.~Kawulok, D.~Kostrzewa, and L.~Skonieczny, ``Evaluating
  super-resolution reconstruction of satellite images,'' \emph{Acta
  Astronautica}, vol. 153, pp. 15--25, 2018.

\bibitem{ChavezRoman2014}
H.~Chavez-Roman and V.~Ponomaryov, ``Super resolution image generation using
  wavelet domain interpolation with edge extraction via a sparse
  representation,'' \emph{IEEE Geoscience and Remote Sensing Letters}, vol.~11,
  no.~10, pp. 1777--1781, Oct 2014.

\bibitem{Demirel2011}
H.~Demirel and G.~Anbarjafari, ``Discrete wavelet transform-based satellite
  image resolution enhancement,'' \emph{IEEE Trans. on Geoscience and Remote
  Sensing}, vol.~49, no.~6, pp. 1997--2004, 2011.

\bibitem{Dong2014}
C.~Dong, C.~C. Loy, K.~He, and X.~Tang, ``Learning a deep convolutional network
  for image super-resolution,'' in \emph{Proc. ECCV}.\hskip 1em plus 0.5em
  minus 0.4em\relax Springer, 2014, pp. 184--199.

\bibitem{Dong2016a}
C.~Dong, C.~C. Loy, and X.~Tang, ``Accelerating the super-resolution
  convolutional neural network,'' in \emph{Proc. ECCV}.\hskip 1em plus 0.5em
  minus 0.4em\relax Springer, 2016, pp. 391--407.

\bibitem{Ducournau2016}
A.~Ducournau and R.~Fablet, ``Deep learning for ocean remote sensing: An
  application of convolutional neural networks for super-resolution on
  satellite-derived {SST} data,'' in \emph{Proc. WPRRS}, 2016, pp. 1--6.

\bibitem{Farsiu2004}
S.~Farsiu, M.~D. Robinson, M.~Elad, and P.~Milanfar, ``Fast and robust
  multiframe super resolution,'' \emph{IEEE Trans. on Image Process.}, vol.~13,
  no.~10, pp. 1327--1344, 2004.

\bibitem{Hardie2007}
R.~Hardie, ``A fast image super-resolution algorithm using an adaptive wiener
  filter,'' \emph{IEEE Trans. on Image Process.}, vol.~16, no.~12, pp.
  2953--2964, 2007.

\bibitem{Irani1991}
M.~Irani and S.~Peleg, ``Improving resolution by image registration,''
  \emph{CVGIP: Graphical models and image processing}, vol.~53, no.~3, pp.
  231--239, 1991.

\bibitem{Kawulok2018Gecco}
M.~Kawulok, P.~Benecki, D.~Kostrzewa, and L.~Skonieczny, ``Evolving imaging
  model for super-resolution reconstruction,'' in \emph{Proc GECCO}.\hskip 1em
  plus 0.5em minus 0.4em\relax New York, NY, USA: ACM, 2018, pp. 284--285.

\bibitem{Kawulok2018EvoStar}
------, ``Towards evolutionary super-resolution,'' in \emph{Applications of
  Evolutionary Computation}, K.~Sim and P.~Kaufmann, Eds.\hskip 1em plus 0.5em
  minus 0.4em\relax Cham: Springer International Publishing, 2018, pp.
  480--496.

\bibitem{Kim2016}
J.~Kim, J.~Kwon~Lee, and K.~Mu~Lee, ``Accurate image super-resolution using
  very deep convolutional networks,'' in \emph{Proc. IEEE CVPR}, 2016, pp.
  1646--1654.

\bibitem{Lai2017}
W.-S. {Lai}, J.-B. {Huang}, N.~{Ahuja}, and M.-H. {Yang}, ``{Fast and Accurate
  Image Super-Resolution with Deep Laplacian Pyramid Networks},'' \emph{ArXiv
  e-prints}, Oct. 2017.

\bibitem{Ledig2017}
C.~Ledig, L.~Theis, F.~Husz{\'a}r, J.~Caballero \emph{et~al.},
  ``Photo-realistic single image super-resolution using a generative
  adversarial network.'' in \emph{Proc. CVPR}, vol.~2, no.~3, 2017, p.~4.

\bibitem{LiJia2008}
F.~Li, X.~Jia, and D.~Fraser, ``Universal {HMT} based super resolution for
  remote sensing images,'' in \emph{Proc. IEEE ICIP}, 2008, pp. 333--336.

\bibitem{Liebel2016}
L.~Liebel and M.~K{\"o}rner, ``Single-image super resolution for multispectral
  remote sensing data using convolutional neural networks,'' in \emph{Proc.
  ISPRSC}, 2016, pp. 883--890.

\bibitem{Liu2016}
D.~Liu, Z.~Wang, B.~Wen, J.~Yang, W.~Han, and T.~S. Huang, ``Robust single
  image super-resolution via deep networks with sparse prior,'' \emph{IEEE
  Trans. on Image Process.}, vol.~25, no.~7, pp. 3194--3207, 2016.

\bibitem{Schultz1996}
R.~R. Schultz and R.~L. Stevenson, ``Extraction of high-resolution frames from
  video sequences,'' \emph{IEEE Trans. on Image Process.}, vol.~5, no.~6, pp.
  996--1011, 1996.

\bibitem{Zhu2016}
H.~Zhu, W.~Song, H.~Tan, J.~Wang, and D.~Jia, ``Super resolution reconstruction
  based on adaptive detail enhancement for {ZY-3} satellite images,''
  \emph{ISPRS Annals of Photogrammetry, Remote Sensing and Spatial Information
  Sciences}, pp. 213--217, 2016.

\end{thebibliography}
\end{document}